# Transforming the Voice of the Customer:
# Large Language Models for Identifying Customer Needs

by

Artem Timoshenko, Chengfeng Mao, and John R. Hauser

January 2026

Artem Timoshenko is an Associate Professor of Marketing at Kellogg School of Management, Northwestern University, 2211 Campus Drive, Suite 5391, Evanston, IL 60208, (617) 803-5630, artem.timoshenko@northwestern.edu.

Chengfeng Mao is a PhD Student at the MIT Management School, Massachusetts Institute of Technology, E62-366, 77 Massachusetts Avenue, Cambridge, MA 02139, (217) 281-2220, maoc@mit.edu.

John R. Hauser is the Kirin Professor of Marketing, MIT Management School, Massachusetts Institute of Technology, E62-538, 77 Massachusetts Avenue, Cambridge, MA 02139, (617) 253-2929, hauser@mit.edu.

We thank Carmel Dibner, Kristyn Corrigan, John Mitchell, and Maggie Hamilton at Applied Marketing Science, Inc. for insightful discussions and research collaboration. We thank Shitong Xiao for outstanding research assistance. This paper has benefited from presentations at the 2024 Emory Marketing Camp, 2024 Symposium on AI in Marketing, 2024 Insights Association Annual Conference, 39th ISMS Marketing Science Conference, 2025 AIML Conference, Wharton's 3rd Annual Business & GenAI Conference, 2nd Open and User Innovation Conference, 2026 MSI Forum, and research seminars at University of Texas at Dallas, Colorado University at Boulder, and the University of Florida.

# Transforming the Voice of the Customer:
# Large Language Models for Identifying Customer Needs

## Abstract

Identifying customer needs (CNs) is fundamental to product innovation and marketing strategy. Yet for over thirty years, Voice-of-the-Customer (VOC) applications have relied on professional analysts to manually interpret qualitative data and formulate "jobs to be done." This task is cognitively demanding, time-consuming, and difficult to scale. While current practice uses machine learning to screen content, the critical final step of precisely formulating CNs relies on expert human judgment. We conduct a series of studies with market research professionals to evaluate whether Large Language Models (LLMs) can automate CN abstraction. Across various product and service categories, we demonstrate that supervised fine-tuned (SFT) LLMs perform at least as well as professional analysts and substantially better than foundational LLMs. These results generalize to alternative foundational LLMs and require relatively "small" models. The abstracted CNs are well-formulated, sufficiently specific to guide innovation, and grounded in source content without hallucination. Our analysis suggests that SFT training enables LLMs to learn the underlying syntactic and semantic conventions of professional CN formulation rather than relying on memorized CNs. Automation of tedious tasks transforms the VOC approach by enabling the discovery of high-leverage insights at scale and by refocusing analysts on higher-value-added tasks.

*Keywords:* Voice of the Customer, Customer Needs, Marketing Research, Product Development, Innovation, Machine Learning, Generative AI, Large Language Models

# Introduction

Understanding customer needs (CNs) is essential for effective product development, product management, and marketing strategy. For example, a snowplow company recognized that when customers plow sidewalks, they often need to turn from one narrow perpendicular sidewalk to another. This insight motivated them to develop a zero-turn snowplow brand that immediately solved an important CN. The movie theater industry was revolutionized with stadium seating to fulfill the CNs of a clean and unobstructed view, room to rest arms, spacious well-cushioned seats, easy to get in and out, rest my head, and storage for refreshments. The identification of patient and physician CNs, such as easy-to-interpret diagnostic information, convenient-sized output, and easy-to-hold, led to a breakthrough medical device (Hauser 1993).

CNs are natural language statements that reveal desired underlying benefits, "jobs to be done." They are articulated at an abstract level to indicate what the product seeks to accomplish, rather than which specific attributes fulfill a CN: clean, unobstructed view in a movie theatre, rather than a 26" wide, swivel chair with cupholders positioned three feet from the next row. Breadth is important, missing important CNs often means the difference between success and failure. Professional studies aim to exhaustively capture CNs that are prioritized using customer surveys and organized in "affinity diagrams" to help product managers focus (Griffin and Hauser 1993).

Identifying CNs is a time-consuming and cognitively demanding manual task. Because customers rarely articulate CNs explicitly, professional analysts must interpret raw qualitative data, such as interview transcripts and online reviews, to distill precise, nuanced insights. Interpretation and formulation require both skill and patience; analysts are trained to develop a deep understanding of the customer’s experience to recognize and articulate CNs accurately. This manual approach is expensive, difficult to scale, and slows the time-to-market.

The abstract and context-dependent nature of CNs means that automating CN abstraction is challenging. While keyword searches and machine learning help to screen large datasets, the critical final step of formulating precise CNs still relies on human expertise (Timoshenko and Hauser 2019). If this tedious task were automated, VOC studies would become faster and more comprehensive. Professional analysts could focus more time on improvements to products and marketing.

Large language models (LLMs) show promise, yet their ability to formulate CNs for professional applications remains an open question. LLMs write text well and have achieved success in abstractive summarization tasks (e.g., Arora, Chakraborty, and Nishimura 2025). However, industry adoption requires precision in capturing the CNs. Poorly formulated CNs, even if automated, would be counter-productive because they focus firms on the wrong sets of questions and do not provide the nuance of “jobs to be done.”

While prompt engineering can generate CNs that "sound reasonable," these outputs often lack depth, nuance, and reliability (see related research by Gao et al. 2024). Supervised fine-tuning (SFT), in which the LLM’s parameters are calibrated with professionally sourced examples, addresses these concerns by mimicking how professional analysts learn to formulate CNs from source material (Dong et al. 2023, Lewis et al. 2020). The efficacy of SFT is often bottlenecked by data accessibility because professional VOC studies are typically proprietary and the cognitive complexity of CN abstraction remains beyond the reach of untrained crowdsourced workers.

Evaluation is equally challenging. We must assess abstracted CNs with respect to deep insights rather than superficial language. For example, the wood-stain-product CN of "able to achieve a desired finish (e.g., satin, semi-gloss, gloss)" might be expressed by customers as "a product that can give my wood an aged look," "assured the final result is not cloudy," "can achieve a glossy or flat finish, depending on my preference" and other phrases. Whether or not these phrases represent well-articulated CNs, provide the necessary specificity, and are true to the source material requires professional judgement. Ground truth is hard to obtain, in part because LLMs are trained to provide statements that sound right to the untrained ear (Ji 2024, Selinger 2024).

Working with an industry partner with over 35 years of VOC experience, we obtained professionally formulated CNs to fine-tune LLMs, and conducted (blinded) professional evaluations to evaluate LLMs' ability to abstract CNs. Our paper describes five interrelated studies that focus on different product or service categories and vary in methods. We show that SFT LLMs identify CNs at least as well as professional analysts and substantially better than foundational LLMs. We examine whether the foundational LLM matters, establish guidelines for the amount of training data for fine-tuning, and investigate the required model size (number of parameters). We show that advanced few-shot prompting is substantially better than zero-shot prompting, but not as good as an SFT LLM in the tested product category.

We look "under the hood" to establish that the SFT LLM does not memorize CNs, but rather efficiently abstracts a given source's content. Fine-tuning uses examples from past VOC applications to modify the LLM's parameters, enabling the model to execute complex syntactic and semantic transformations necessary to abstract CNs, rather than solutions, targets, and opinions. To understand the SFT LLM in the field, we describe key managerial insights for one application and summarize four other applications. SFT LLMs are transforming VOC applications by automating tedious, expensive, and time-consuming tasks, thus freeing up professional analysts to focus on higher-value-added tasks in innovation. The ability to scale to more source material means that more high-leverage CNs are discovered, leading to better applications.

## Related Literature

### *Identifying Customer Needs (CNs)*

CNs are the basis of the voice-of-the-customer (VOC, Griffin and Hauser 1993; hereafter, GH 1993). In the last 30+ years, there have been hundreds of academic and industry articles on improved methods for qualitative interviews, ethnographic methods, metaphor elicitation, and interpretation (e.g., Brown and Eisenhardt 1995, Burchill and Brody 1997, Gupta 2020, Mitchell 2016, Zaltman 1997, Cayla, Beers, and Arnould 2014). All proposed methods require human judgment to interpret customer interviews.

More recently, firms recognized that user-generated content (UGC, e.g., online reviews, blogs, and forums) augments customer interviews, thus requiring new methods that scale CN analysis to large-volume UGC. Initial research focused on the word counts, word co-occurrence, and topic models to identify "bags of words," but "bags of words" do not describe nuanced CNs (Lee and Bradlow 2011, Netzer et al. 2012, Schweidel and Moe 2014, Büschken and Allenby 2016). To help identify more-nuanced CNs, Timoshenko and Hauser (2019; hereafter, TH 2019) use convolutional neural networks to identify diverse informative sentences to be reviewed by professional analysts to abstract CNs. Our paper investigates whether LLMs can successfully automate this last human-centric step to identify high-quality CNs.

***Large Language Models***

Large Language Models (LLMs) use deep and parallel neural network layers and are pretrained on vast amounts of text data to understand, generate, and respond to natural human language. Current LLMs, such as GPT-5, Claude, and LLaMA 4, combine transformer architecture (Achiam et al. 2023, Touvron et al. 2023a) and feedforward layers (multilevel perceptron layers). The self-attention modules in transformers handle sequential data at scale, allowing for parallel processing and capturing long-range dependencies in text (Vaswani et al. 2017). The perceptron layers mostly store factual information and make up about two-thirds of the parameters (Geva et al. 2021). LLMs are typically trained using a combination of self-supervised learning and reinforcement learning from human feedback (Christiano et al. 2017).

LLMs have demonstrated remarkable capabilities across domains, such as text summarization (Ibrahim 2025), education (Kasneci et al. 2023, Lo 2023), healthcare (Moor et al. 2023, Thirunavukarasu et al. 2023, van Veen et al. 2024), coding (Gao et al. 2023), financial services (Yang et al 2020), and law (Deroy et al. 2024, Katz et al. 2024). Many of these applications focus on summary rather than transformation (such as abstracting CNs from source material). The marketing science community has explored applications of LLMs for marketing research and practice. One prominent idea is that LLMs can serve as synthetic respondents (Horton 2023). For example, Arora, Chakraborty, and Nishimura (2025) use LLMs to create marketing personas that answer qualitative and quantitative questions. Brand, Israeli, and Ngwe (2023), Wang, Zhang, and Zhang (2024), and Eggers and Vriens (2026) integrate LLMs into conjoint studies to obtain more-precise preference measures. Qiu, Singh, and Srinivasan (2023) evaluate LLMs for eliciting consumer risk preferences. Li et al. (2024) explore LLMs for automated perceptual mapping. Dong (2024) uses LLMs to successfully replicate the customer decision rules identified by human judges in unstructured direct elicitation (Ding et al. 2011).

LLMs do well on many tasks, but not all tasks. Gao et al. (2024, p. 2) suggest that many LLMs "differ markedly from that of human participants" and "exhibit unstable behavior that differs from human behavior to a statistically significant degree, regardless of the approach used." This variation is particularly prominent in new tasks that the LLM has not memorized from its vast training data. Prompt engineering is often effective (Brown 2020), but it does not always work. It is a challenge to consistently generate high-quality prompts (Min et al. 2022). Not surprisingly, academic research and industry practitioners report challenges with the stability of wrapper applications when foundational LLMs introduce updates (Chen, Zaharia, and Zou 2024). Challapally et al. (2025) suggest that 95% of industry applications do not achieve their targets. Lu et al. (2022) demonstrate that examples provided in prompt engineering are not always effective; differing orders of prompts can result in either excellent or random-guess performance. For human-oriented decisions, Gao et al. (2024) demonstrate that simple queries, prompt engineering, and providing external documents as references (retrieval automated generation, RAG) differ in distribution from human respondents. We add to this literature by providing a careful evaluation of a specific, but managerially important application of LLMs.

## Industry Practice

Before we evaluate whether an SFT LLM can be used in industry to abstract CNs, we need to understand the professional analysts' task in the VOC studies.

Formulating CNs is demanding. The challenge lies in understanding the deeper motivations that drive customer behavior and capturing these motivations in a concise and efficient form. Industry professionals often differentiate between CNs, solutions, targets, and opinions. For example, a customer might express dissatisfaction with battery life in cellphones (an opinion), but the underlying CN might be the desire for longer, uninterrupted use while traveling. In academic literature, solutions and targets are often framed as product attributes, while opinions reflect customer sentiments.

Understanding well-formulated CNs before focusing on specific solutions provides insights beyond the current market offerings. Before the zero-turn snowplow was launched, there was no snowplow that could move from one perpendicular sidewalk to another. The ability to do so was not a defined attribute. Before stadium seating was introduced to movie theaters, attributes such as drink holders and elevation were not part of the conversation. A professional service organization may wish to attract new members, but not recognize that new members must learn the specialized informal jargon before they can benefit from conferences, training, and certification.

To identify CNs, firms use experiential interviews, metaphor elicitation, ethnography, focus groups, call center logs, and/or user-generated content to create a corpus of sentences and phrases. Professional analysts highlight relevant sentences and phrases in the source material and interpret the customers' words as CNs, keeping the verbiage and the customer's intentions as close as feasible to those articulated by the customer. Customers rarely articulate CNs clearly, so the analysts' task involves interpretation (abstraction) of the underlying CNs, rather than merely paraphrasing.

**Example 1**. A customer complained about a 30-second timer in a toothbrush: "*I replaced an old brush with a new one, BUT the description doesn't say that this model no longer has a 30-second timer. The brush shuts off after 2 minutes but the 30 second timer is missing. I would not have purchased this product if I had known.*" From this review, a professional analyst abstracted a CN

"*Able to know the right amount of time to spend on each step of my oral care routine.*"

CNs tend to be nuanced, yet not so specific that they forestall creativity. Overly generic CNs, such as "*ease of use*," fail to provide actionable insights and meaningful ideas for innovation. Conversely, too-specific CNs, such as "*able to dry in 20 minutes at 70% humidity*" limit creativity and fail to generalize.

**Example 2**. A professional study identified three CNs focused on breath freshness: "*Able to eat and drink anything and my breath still stays fresh*," "*Able to have fresh breath all day, i.e., no need to keep freshening it*," and "*Able to tell if I have bad breath*." Each CN identifies a specific "job to be done," all related to a more-general area of breath freshness. If the product developer or marketer can identify ways to satisfy these CNs, the customer might pay a premium for the solution(s).

Abstracting CNs is a cognitively demanding task, and humans are fallible. GH 1993 report that each analyst was able to identify 54% of the CNs (range 45-68%) that were ultimately identified. With more applications, professional analysts have become more skilled, but still not perfect. New analysts receive training materials that define CNs, contrast CNs with (existing) solutions, and provide standards to abstract CNs. They learn best practices such as formulating CNs as concise positive statements using simple, accessible language that captures the core customer benefit without ambiguity. Analysts are given many examples of CNs, including statements that are not CNs. The analysts hone their craft by formulating CNs and receiving feedback from more experienced colleagues. We attempt to replicate the spirit of this training with example-based fine-tuning.

Depending on the product (or service) category and the source material, analysts might identify hundreds or even thousands of potential CNs. Not surprisingly, there is redundancy.

Using keyword matching and experience-based judgment, analysts “winnow” the identified needs to a less-redundant set that typically consists of 80-120 CNs. Next, to focus product development, product management, and marketing on creative solutions, analysts organize the identified CNs into a three-level hierarchy of primary, secondary, and tertiary CNs (Burchill and Brody 1997, GH 1993). In our paper, the term “customer needs” refers to the tertiary needs in the hierarchy. These CNs are most nuanced and are the critical step in the VOC process.

The winnowing and affinitization tasks require training and experience to channel customers as informed by the source material, but affinitization provides a measure of breadth. An important criterion for evaluating the effectiveness of VOC studies is the ability to identify (tertiary) CNs in each of the primary and secondary levels of the hierarchy – the levels that are most often considered in product development, product management, and marketing strategy discussions.

## Large Language Models for Extracting Customer Needs

From a computer science perspective, formulating CNs requires abstractive summarization. Abstractive summarization involves generating new phrases that capture the core meaning of an input in a more conceptual and concise manner. For instance, an abstractive summary of a news article might condense complex details into "*World leaders discussed global strategies to mitigate climate change and reduce emissions*," even if those exact words do not appear in the original text. In contrast, extractive summarization focuses on identifying and reproducing key phrases, such as when a search engine extracts snippets containing exact sentences from documents. LLMs are well-suited to the more-challenging task of abstractive summarization.

To examine whether LLMs can formulate deep nuanced CNs, we develop prompts for a foundational LLM and gather training data for an SFT LLM. We then recruit new-to-the-study (blinded) professional analysts as professional evaluators of the CNs produced by the LLMs and the original VOC application.

***Foundational Model: Vicuna 13B***

Our primary analysis uses Vicuna 13B as a foundational LLM. "13B" indicates the size of the model – roughly the number of parameters. In our preliminary investigation, CNs formulated by Vicuna 13B were qualitatively similar to the ones formulated by the contemporary state-of-the-art publicly available models (including ChatGPT-4o). We wanted a strong baseline. There was no noticeable improvement for the larger Vicuna 33B. Importantly, Vicuna 13B offers a license for academic use with open weights, which enables reproducibility.

Vicuna is a general-purpose foundational LLM, developed by academic researchers from UC Berkeley, UCSD, and CMU (Chiang et al. 2023). Vicuna uses LLaMA 2 as a base model and improves it with 70,000 user-shared ChatGPT conversations (Touvron et al. 2023b). Vicuna performed comparable to the contemporary open-source and closed-source LLMs in writing tasks (Zheng et al. 2024, Zhang et al. 2023) and has been applied successfully in downstream applications and research (Zhu et al. 2023, Mullappilly 2023).

For our baseline, we sought a strong set of prompts for the foundational LLM. We explored multiple prompt variations, starting with "*Extract customer needs from <Review Text>*," and then adding (1) a definition of a CN, (2) a requirement to formulate a response in a single sentence, (3) examples of CNs (in-context learning), and many other variations and combinations. In exploratory preliminary analysis, all options performed similarly. Overall, the best prompt was: "*For a <Product Category>, identify a customer need from the user review. If* *no need is found, return []. Review: <Review Text>*".

Because our research requires professional evaluators to read and evaluate model outputs, an expensive and time-consuming process (Study 1), we needed to commit to a model architecture and a specific prompt early in the research process. In Study 2, we use the LLM-as-a-Judge approach to explore alternative foundational LLMs and prompting approaches (Gu et al. 2025; Zhang et al. 2025). Evaluation by a trained classifier sacrifices some external validity, but is the only feasible method to compare variations in the foundational LLM architectures, size of the LLMs, amount of training data, and prompting. A large number of such evaluations would be prohibitively expensive if highly-compensated professional evaluators were asked to judge all variations.

***Fine-tuning an LLM with Professional VOC Studies***

We trained an SFT LLM to abstract CNs using examples from professional VOC studies. The training, which mimics professional-analyst training, directly updates the weights of the foundational model using back propagation. Fine-tuning seeks to improve the LLM's performance in a specific application (CN abstraction) and teaches a model to output in an appropriate style and tone. See Figure 1.

**Figure 1**. Fine-tuning an LLM with CNs from Professional VOC Studies

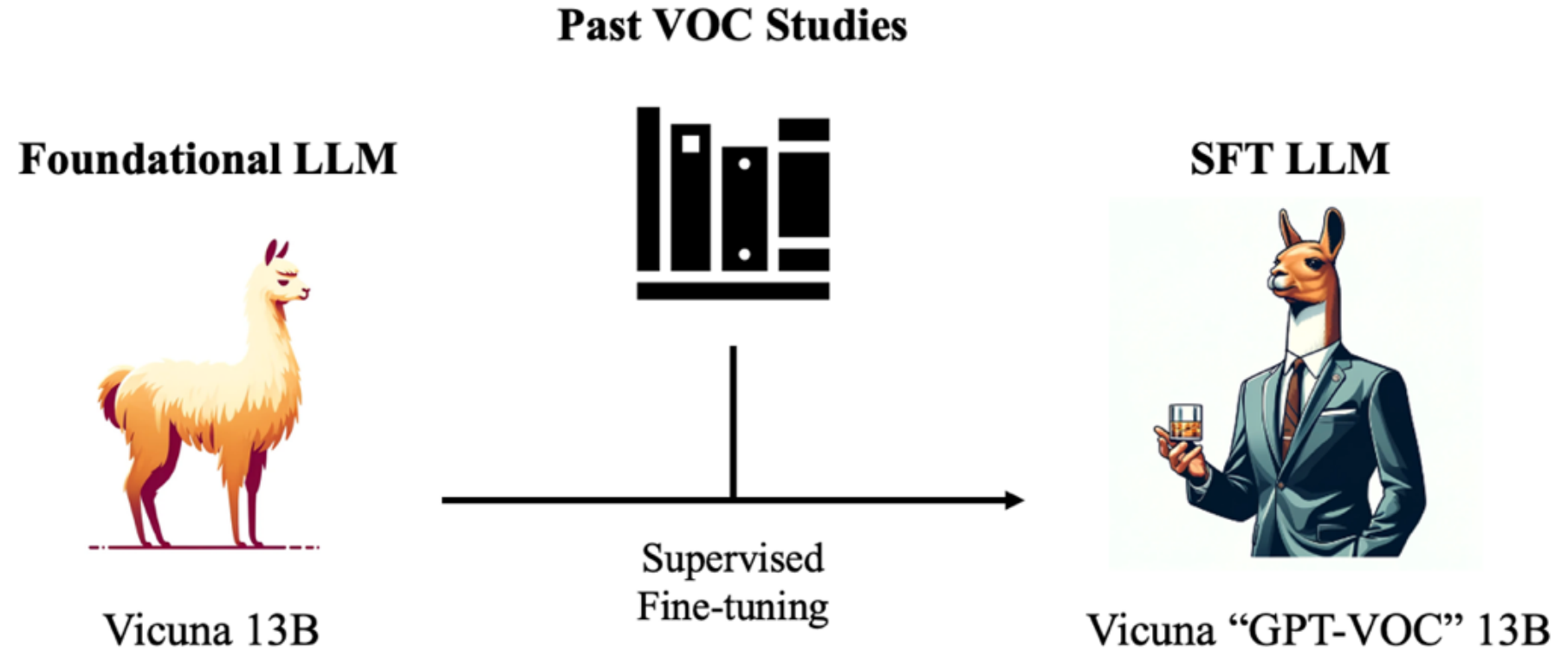

Our fine-tuning data span ten different product categories, such as activewear, glucose monitors, and recreational vehicles. All categories in our training data are physical products – we report on an industry application to services in Study 5. The data from each product category came from VOC studies by a firm with over 35 years of experience and hundreds of successful applications in product development, product management, and marketing. Each application provided valuable insights to a client who either improved their products or improved their marketing. Appendix A provides a complete list of product categories. For each category, the firm reviewed interview transcripts and/or online reviews and identified CNs. Typically, firms have no incentives to retain source material (interviews, online reviews) and rarely are able to match CNs to source sentences. Our data include matched verbatims. We augment these data with "negative" examples that include sentences from interviews and online reviews that contain no information about CNs (uninformative sentences).

**Table 1**. Examples of Training Data for Fine-tuning an LLM

| Question | Answer |
|---|---|
| <GPT-VOC> <Product Category = "**Activewear**"><br><br>Just really curious why Oxford Gray on this is a different color than the Oxford Gray on the powerblend sweats. | Confident that colors will be consistent across products |
| <GPT-VOC> <Product Category = "**Recreational Vehicles**"><br><br>I tested it and it worked really well. | [] |

Table 1 illustrates a positive and a negative example from our training data. The question-and-answer structure is similar to the instruction fine-tuning in natural language processing (Chung et. al 2024). The data provide the LLM with examples of ideal answers to application-specific prompts. We add a special tag <GPT-VOC> and indicate the product category at the beginning of each prompt to condition the model on identifying CNs for the category. Special tags are a common practice to condition the LLM with a new task, such as using augmented tools (Schick et al. 2024) or predicting protein (Shen et al. 2024). The special tag should be unique, but the tag is not required to be meaningful. After fine-tuning, we use the standardized prompt as a shorthand to indicate to the SFT LLM that the task is to abstract CNs.

During fine-tuning, the SFT LLM attempts to abstract CNs and receives feedback on the "correct" output – either the CN or []. We use backpropagation to fine-tune the parameters of the model (Dong et al. 2023). Fine-tuning was facilitated by the DeepSeek Library and required approximately eight hours using four Nvidia A100 GPUs from Lambda Labs – an on-demand AI developer (GPU) cloud service. After calibration, applications are run on a local workstation (desktop).

The ten professional VOC studies provided 1,549 CN-verbatim pairs. We randomly split the "positive" examples into two subsamples for model training (80%) and validation (20%). Additionally, the data contained 11,975 uninformative sentences. We need "negative" examples to train the model to output "[]" for uninformative content, but we must choose the number of negative examples carefully to control the tradeoff between false negatives and hallucination. Too many negative examples lead to missing CNs (false negatives), while too few negative examples lead to CNs that do not follow from the verbatims (hallucination). We selected the number of randomly-chosen negative examples (47) by observing the model performance on the validation data.

The following illustrative example is based on wood-stain products – a category not used in fine-tuning. Studies 1-5 evaluate the SFT LLM using out-of-sample product categories excluded from fine-tuning.

***Illustrative Example of LLM-Based CNs***

Figure 2 illustrates the output of the LLMs for wood stain products. From an online review (source content), professional analysts identified the CN: “*Able to see what surface areas I have already covered*.” This CN seems relevant when wood stain is applied to larger surface areas.

**Figure 2**. Illustrative Example Customer Needs Identified from Online Reviews

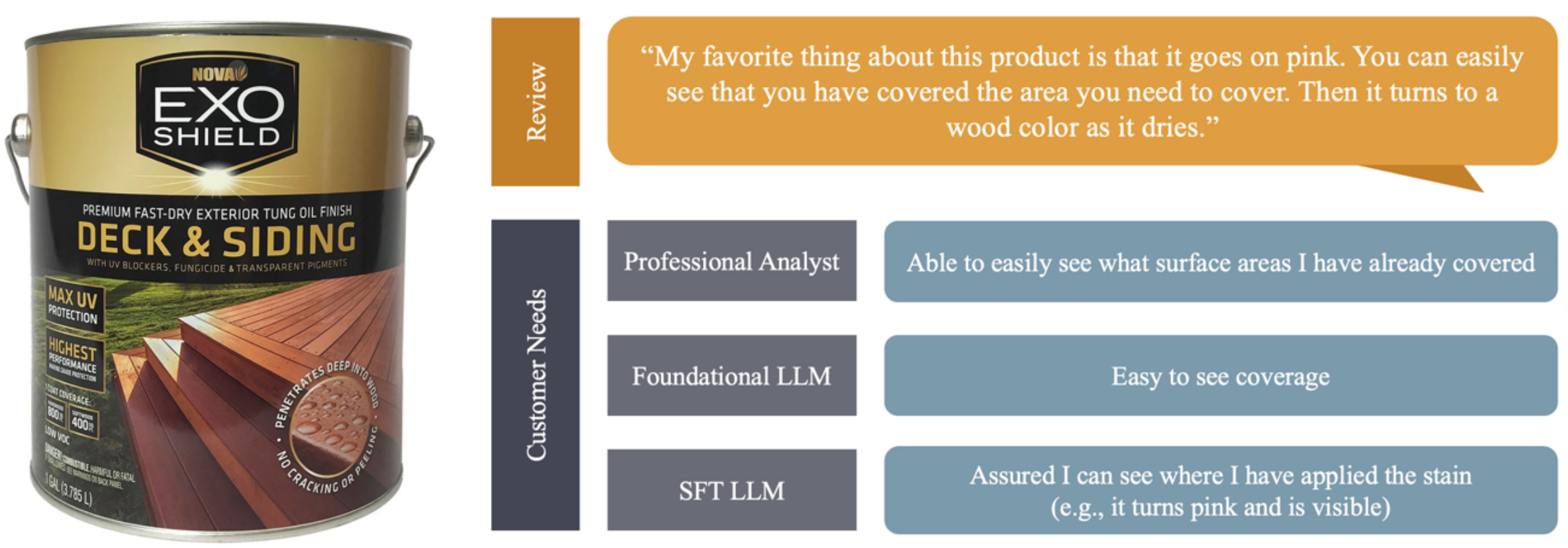

Without fine-tuning, the foundational LLM (Vicuna 13B) abstracted a CN “*Easy to see coverage*.” While this statement correctly summarizes the topic of the online review, the statement lacks the specificity required to guide innovation. The performance of the foundational LLM in Figure 3 is typical; the foundational LLMs often paraphrase customer reviews or elicit solutions and opinions.

The SFT LLM formulated a CN “*Assured that I can see where I have applied the stain*.” This CN captures a “job to be done” from the original content, is concise, and provides sufficient detail for product development. The formulation by the SFT LLM includes a clarification (“*e.g., it turns pink and is visible*”). Our training data contains similar clarifications in 34% of the examples. Appendix C provides additional examples of online reviews and the corresponding CNs as abstracted by professional analysts, the foundational LLM, and the SFT LLM. While the CNs abstracted by the SFT LLM are different from the CNs abstracted by professional analysts, these CNs capture similar meaning and adhere to similar professional standards—a judgment we examine formally in this paper.

## Empirical Evaluation: Study Overview

We evaluate the LLMs' ability to abstract CNs using five interrelated studies. Table 2 provides an overview. The studies vary in research questions, methodology, data sources, and industries.

Our primary evaluations (Studies 1, 2, & 3) focus on wood stain products. Study 1 provides the core evidence comparing CNs formulated by LLMs to the professionally formulated CNs in a blind test. Study 2 explores the limits of the technology, evaluating different foundational models and determining how much training data are required for the SFT approach to maintain high-quality outputs. Study 3 investigates what LLMs learn during fine-tuning with past VOC studies. We contrast the transformation vs. memorization mechanisms, and cross-evaluate failure modes for the foundational and SFT models.

Studies 4 & 5 provide evidence from other industries to establish cross-category generalizability. Study 4 reports an additional blind study with professional analysts in the oral care product category (toothbrushes/toothpaste). This study is consistent with our primary findings. Study 5 evaluates CNs abstracted by the SFT LLM in a service category using interview-based data. We discuss the rich managerial insights from this for-client VOC application and indicate the resulting improvements in the service. None of the three categories—wood stain, oral care, and professional services—were included in the fine-tuning data.

**Table 2**. Study Overview

| Study | Product Category | Primary Research Question(s) | Method & Data Source |
|---|---|---|---|
| Study 1 | Wood Stain Products | Do SFT LLMs abstract customer needs as well as professionals? | Blind evaluation by professional experts; online reviews and professional forums. |
| Study 2 | Wood Stain Products | Does the foundational LLM impact performance? How much training data are necessary? | Sensitivity analysis using fine-tuned transformer-based quality classifiers. |
| Study 3 | Wood Stain Products | Does SFT facilitate transformation or memorization? How do failure modes differ from foundational models? | Comparative failure-mode analysis and cross-evaluation via performance probes. |
| Study 4 | Oral Care Products | Do results generalize to other product categories and different data sources? | Blind evaluation by professional experts; Amazon reviews and CNs from prior research. |
| Study 5 | PDMA (Service Category) | Do insights extend to services and experiential interviews? | Professional VOC application of the SFT LLM in the service industry. Short descriptions of four other applications. |

Our research required domain experts to read the stated CNs and judge whether these statements adhere to professional standards (Griffin 2004, also the PDMA's Glossary for New Product Development). Our research partner uses extensive training and peer support to ensure that the definition of CNs is consistent with professional standards. The firm has conducted numerous VOC studies for consumer brands and business-to-business organizations, and is often called upon to train other firms in CN identification. For the blind studies, we recruited professional evaluators from our research partner. The evaluators were not involved in the initial for-client VOC studies with these product categories. Our methodology further assured the evaluators were blind to whether the CNs were formulated by other professional analysts, the foundational LLM, or the SFT LLM.

## Studies 1-3: Professional Analysts vs. LLMs for Wood Stain Products

The empirical analysis in this section builds on a professional VOC study that identified 103

CNs for wood stain products. We augment these data with LLM-based CNs and additional labels to investigate the effectiveness of LLMs for VOC applications.

The original VOC study used a machine-learning approach to screen 14,341 online-review sentences and identify 1,000 informative and non-repetitive sentences to ensure diverse content (TH 2019). The 1,000 number was selected as a typical cost-versus-quality tradeoff for client-based VOC applications. Following standard procedures, the professional analysts read the selected online reviews and manually identified unique CNs. The firm shared with us the verbatims for every unique CN. We applied the foundational and SFT LLMs to identify CNs from the same online reviews.

To minimize information leakage, the wood stain category was not used in LLM fine-tuning. Although some information on wood stains might have been available during the foundational training of Vicuna 13B, professional VOC studies and CN formulations are trade secrets and unlikely to be available publicly. Any leakage in foundational training for Vicuna 13B would only reinforce the key qualitative recommendations in this section.

***Study 1a. Are Statements Abstracted by LLMs Well-Formulated Customer Needs?***

Our first study evaluates whether CNs abstracted by professional analysts and LLMs adhere to the industry's professional standards. Three professional evaluators, not involved in the original wood stain application and with experience in VOC studies, evaluated CN statements on three dimensions. The wording of the questions to the evaluators was based on extensive discussion with the firm's VOC experts. The questionnaire was pretested and revised so that each question was clear, understandable, and measured the target construct. Appendix D provides detailed instructions, including the user interface of the study design and clarifications about the evaluation dimensions. The basic questions asked are paraphrased below.

(1) The statement qualifies as a CN identified in a typical VOC study ("Is Customer Need")

(2) The statement captures sufficient detail about a CN ("Sufficient Detail")

(3) The statement is based on some information in the review ("Follows from the Source")

Each of the three professional evaluators evaluated 150 randomly-chosen sentences from online reviews. For each review sentence, a professional evaluator was given the text of the online review and three CN statements (professional analyst, foundational LLM, and SFT LLM). We randomized the order of CNs for each review. Evaluators were blind to the purpose of the study, and posttests indicated that there were no inadvertent cues about how the CNs were extracted. We aggregated individual evaluations using majority voting. In the following analysis, each data point corresponds to a review-CN combination.

The sample of 150 review sentences included (1) 90 sentences indicated as *verbatims* leading to CNs in the voice-of-the-customer application, (2) 30 reviews indicated by the firm as *informative* but not used as verbatims, and (3) 30 reviews indicated by the firm as *uninformative*. The firm considered the uninformative reviews in the original for-client VOC study and decided that they did not contain CNs. For *verbatims*, all three approaches identified CNs. For the other categories of reviews, the professional analysts in the original VOC study decided not to formulate CNs. To maintain blinding, every question contained three plausible CNs to avoid any inadvertent signals about how a CN was formulated. For non-verbatims, we augmented the CNs from the foundational and SFT LLMs with analyst-identified CNs randomly selected from the original VOC application.[1]

Figure 3 reports results for the first two questions, aggregated across verbatim, informative,

[1] After considering many alternatives ways of choosing analyst-identified CNs, we settled on a strategy of randomly chosen, but real, CNs. This strategy is faithful to the original VOC professional study and consistent with the statistics cited earlier that human analysts do not identify 100% of the CNs.

and uninformative reviews. (Appendix E reports the disaggregated results.) The professional-analyst CNs provide an important benchmark for the LLMs. Professional analysts identified these CNs in a for-client application. Despite extensive training, professional analysts are not perfect. In 1993, Griffin and Hauser reported imperfect (54%) abstraction of CNs from experiential interviews. In Study 1, the professional evaluators agreed with the original professional analysts about 80% of the time on "Is Customer Need" and "Sufficient Detail." Although the task in Figure 3 is not identical to the Griffin-Hauser task, the 80% agreement suggests improvement in industry practice and reinforces our decision to rely on professional evaluators rather than research assistants or untrained crowdsourced workers.

**Figure 3.** Comparison of Customer-Need Abstraction by Professional Analysis and LLMs

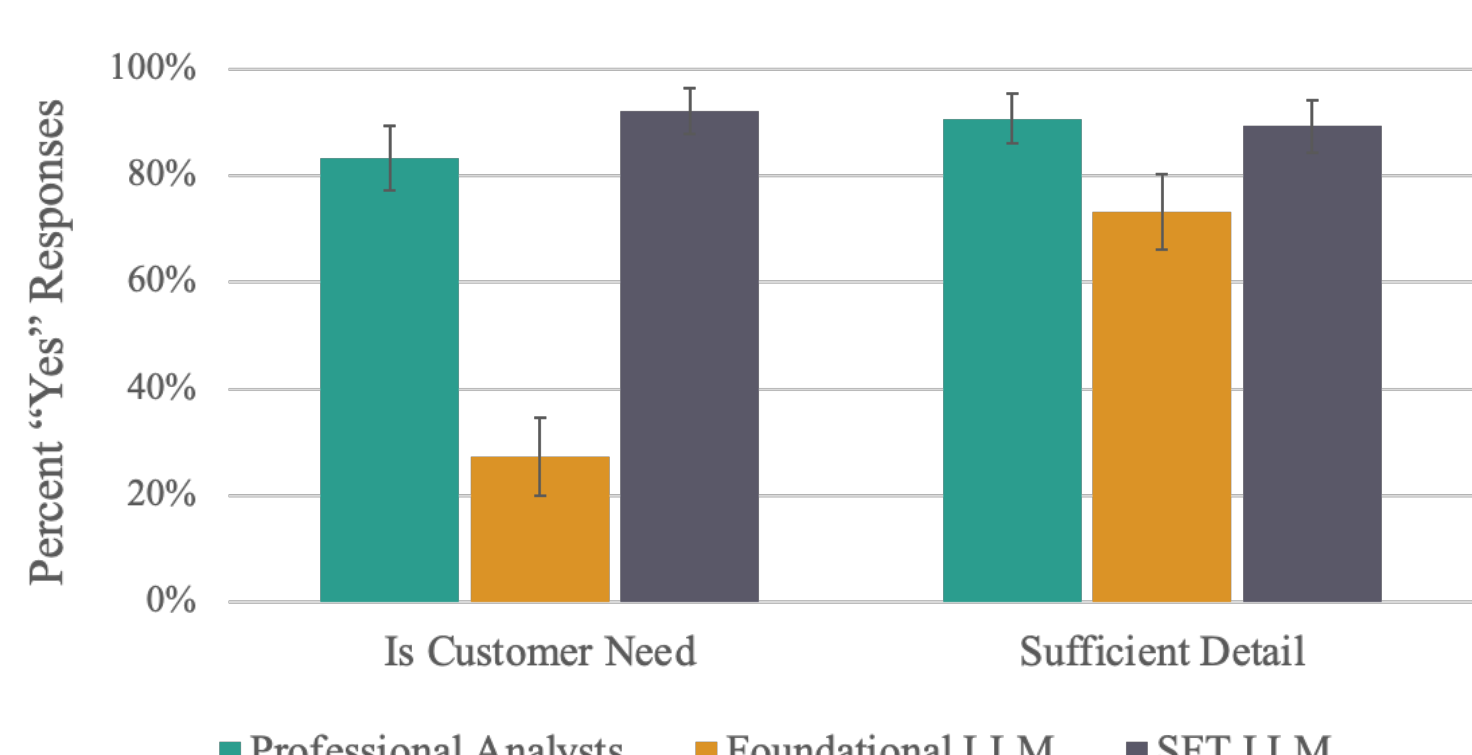

We use a z-test to test the statistical significance of differences between models. The patterns with McNemar's paired-proportions test are consistent. We use Cohen's $h$ to report effect sizes. The error bars in Figures 3-10 represent 95% confidence intervals.

The SFT LLM is effective. The SFT LLM CNs are significantly more likely to be judged as typical-to-VOC CNs than those identified by (fallible) professional analysts, but the effect size is small ($z = 2.29$, $p = 0.02$, Cohen's $h = 0.27$). The SFT LLM CNs also capture as much detail as professional analysts ($z = 0.40, p = 0.69$, $h = 0.04$). These results validate the use of SFT LLMs to identify CNs.

Figure 3 indicates that the foundational LLM is not a viable solution for this product category. The foundational LLM does significantly and substantially worse than both professional analysts ($z = 9.76$, $p < 0.01$, $h = 1.20$) and an SFT LLM when abstracting CNs ($z = 11.42$, $p < 0.01$, $h = 1.47$). The foundational LLM does significantly worse than professional analysts and an SFT LLM when capturing detail, although the effect sizes are moderate ($z = 3.91, p < 0.01, h = 0.47$, and $z = 3.56, p < 0.01, h = 0.42$, respectively). Without fine-tuning, the foundational LLM is not a viable option for identifying CNs.

We next examine whether hallucinations are a problem for either of the LLMs (Rawte et al. 2023) or for professional analysts. Recall that customers do not state CNs directly. Professional analysts and LLMs interpret information in the verbatims to abstract the customer's implied "job to be done." Analysts might hallucinate after tediously reading many source sentences. LLMs might hallucinate based on the extensive data on which the foundational LLM was based.

In Figure 4, we evaluate whether the abstracted CNs follow from the *verbatims* known to contain a CN. The evaluators answered: "Please evaluate whether or not the statement is based on information in the review. In particular, is it reasonable that a VOC study would abstract this customer need from the review?"

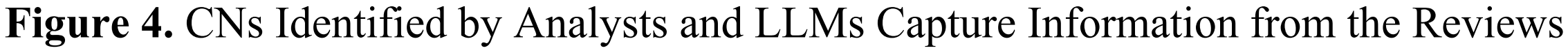

**Figure 4.** CNs Identified by Analysts and LLMs Capture Information from the Reviews

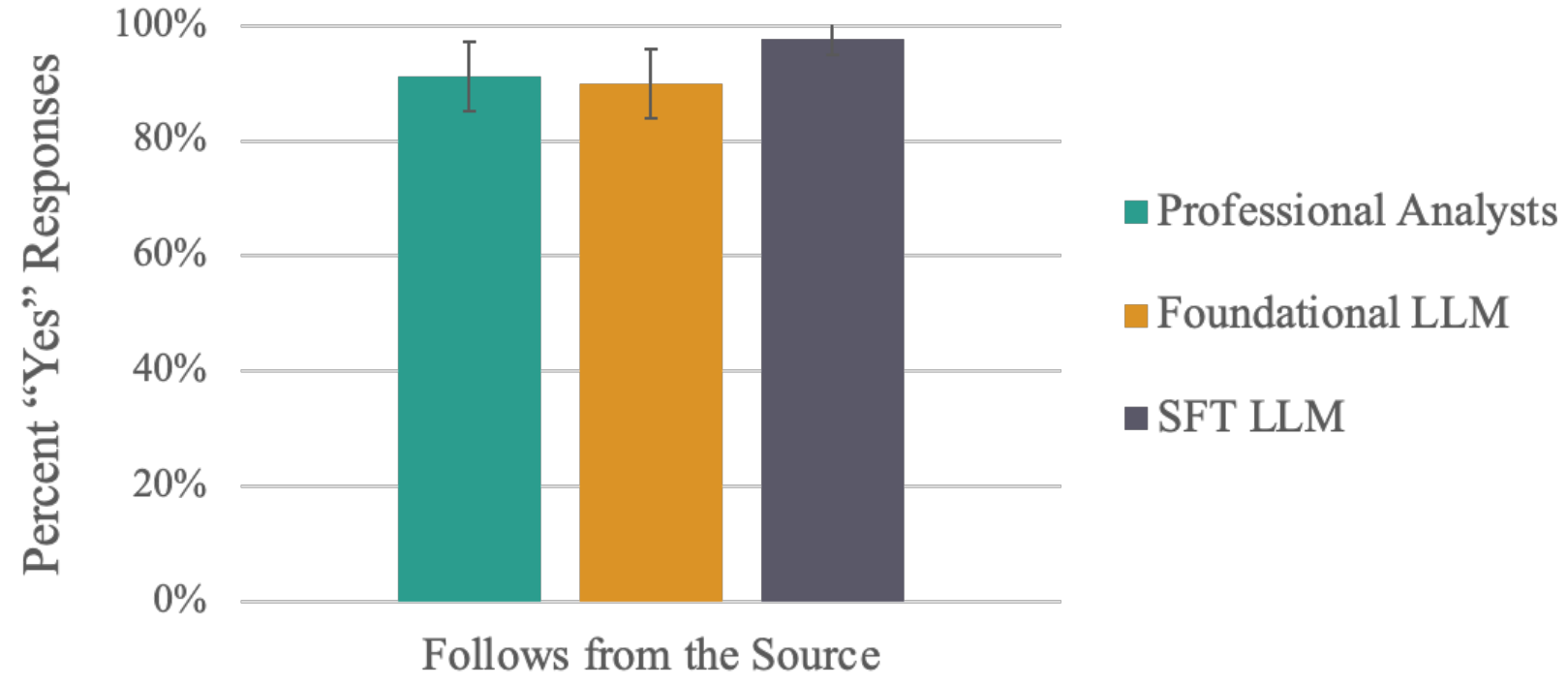

Figure 4 focuses on the 90 *verbatims*. Limiting Figure 4 to verbatims provides a fair comparison among professional analysts and LLMs and, if anything, favors professional analysts. The appendix reports data from all other review sentences. The data are face valid and reinforce our interpretations.

The professional-analyst-abstracted CNs were judged to closely represent information from the verbatims for 91% of observations. The firm reports that this percentage is within an expected range for VOC applications. The SFT LLM significantly outperforms professional analysts although the effect size is modest ($z = 2.53, p = 0.01, h = 0.31$). The foundational LLM performs similarly to the professional analysts ($z = 0.33, p = 0.75, h = 0.04$). Figure 4 suggests that LLM hallucinations are not a practical problem for identifying CNs.

***Study 1b. Can LLMs Capture Diverse Customer Needs?***

Typical VOC applications identify hundreds of CNs from raw qualitative data. To be useful for product development, product management, and marketing, CNs are winnowed to remove duplicates and grouped into an affinity diagram—a hierarchical structure of primary CNs, secondary CNs, and tertiary CNs (Burchill and Brody 1997; GH 1993). Redundancy among CNs is reduced, and higher-level CNs (primary and secondary) are chosen to represent the customer's perspective. For example, for wood stain products, the CN "a*ble to control depth and finish of the stain and topcoat*" could belong to the secondary CN "*application looks even and consistent,*" from the primary CN "*appearance and finish*." Professional experience suggests that applications are most successful when a VOC study identifies CNs in every primary and secondary category. To evaluate whether the LLMs capture a broad set of CNs, we use an affinity diagram based on a concatenation of the professional-analyst and SFT-LLM CNs. Our research partner developed the affinity diagram in three steps.

*Step 1. Preliminary Winnowing*: Winnowing relies on human judgment to eliminate

redundancy, often merging CN-statements at the tertiary level. Our research partner first winnowed the SFT-LLM-identified CNs obtained from the 14,341 source verbatims. Analysts used the same procedure used for professional VOC applications to return a preliminary set of 154 winnowed CNs.

*Step 2. Final Winnowing and Affinitization:* We merged the 154 winnowed SFT-LLM-identified CNs with the 103 analyst-identified CNs from the original VOC application. Experienced analysts, not involved in the original application or in Study 1, constructed an affinity diagram. Following standard practice during the affinitization process, the analysts further winnowed the CNs to a final set of 117 CNs. The analysts were blind to the source of CNs and preserved the mapping from the 257 (154 plus 103) merged CNs to the final 117 winnowed CNs.

*Step 3. Mapping CNs Back to Verbatims:* Our research partner reconstructed the mapping from a randomly-sampled set of 2,000 SFT-LLM-identified CN statements to the final 117 CNs. For each of the 2,000 SFT-LLM-identified statements (pre-winnowing), professional analysts mapped the 2,000 statements to the final 117 CNs. The mapping is many-to-many.

The SFT LLM made the winnowing task in Step 1 substantially less tedious and time-consuming because analysts found it easier to winnow the CNs directly rather than winnowing the source material. In standard practice by professional analysts, it is much more efficient to winnow the source material prior to CN abstraction to avoid the time-consuming task of abstracting CNs from every potential verbatim. Efficiencies in winnowing enable the SFT LLM to scale to much larger source corpora.

The mapping required a major effort over several months, and it would have been cost-prohibitive to reconstruct a mapping for all CNs identified from the entire set of 14,341 source verbatims. In typical VOC applications, the mapping from final CNs to source verbatims is not retained because the mapping is not considered valuable for business applications. The substantial cost of reconstructing the mapping is rarely justified by any corresponding benefits. Indeed, in over thirty-five years of VOC applications by our research partner, they know of only one instance in which the firm recreated the mapping—a litigation-support application which required extensive documentation.

*Strategic Value: Managerially Relevant Primary and Secondary Customer Needs*

For wood-stain products, the final affinity diagram includes 30 secondary CNs which were in turn grouped into eight primary CNs. The CNs identified by the SFT LLM had slightly more strategic breadth. The SFT LLM identified CNs from 100% of the primary groups and 100% of the secondary CN groups. The professional analysts identified CNs from seven (87.5%) of the primary groups and 24 (80%) of the secondary groups. For example, the analysts in the original study missed the primary CN which describes *a desire for products that help make the maintenance of wood easier*.

Looking at the tertiary CNs, the 154 winnowed SFT-LLM-identified CNs account for 84.6% of the final affinitized tertiary CNs while the 103 winnowed analyst-identified CNs account for 48.7%. (The percentages add to more than 100% because of overlap.) These percentages must be interpreted cautiously. Typical cost-vs.-benefit considerations limited the original VOC study to a sample of the corpus, but the SFT-LLM scaled well to the full corpus. Secondly, these percentages confound redundancy reduction (both stages of winnowing) with whether an after-winnowing-and-affinitization-tertiary CN is equivalent to one of the CNs abstracted from the source material.

### *Study 2. Generalizability and Boundary Conditions*

Studies 1a and 1b demonstrate that a fine-tuned LLM (Vicuna 13B) performs on par with professional analysts in CN abstraction. Study 2 explores the boundaries of this performance by testing the impact of the amount of training data, model size, and generalizability across different foundational models. Additionally, we evaluate whether few-shot prompting bridges the gap between foundational and fine-tuned models.

High-cost professional-analyst evaluation is infeasible for these many variations. To address this, we adopt an "LLM-as-a-judge" approach (Gu et al. 2025; Li et al. 2025), using expert ratings from Study 1 to train a transformer-based quality classifier (henceforth, “the classifier”). This classifier replicates professional-analyst (human) judgment and allows us to evaluate CN statements generated by different models at scale.

Our approach is motivated by recent literature. Research suggests that LLMs are scalable and particularly good at open-ended evaluations, but without additional calibration, may be biased (Shi et al. 2025; Panickssery et al. 2024). For focused yes/no tasks, pretrained transformer-based quality classifiers are particularly effective; they significantly outperform zero-shot LLMs on specific text classification when sufficient labeled data are available for fine-tuning (Bucher and Martini 2024; Chae and Davidson 2025; Zhang et al. 2025).

The classifier’s task is fundamentally simpler than the SFT LLM’s task. The SFT LLM performs the generative task of abstracting nuanced CNs from source material, whereas the classifier performs a binary classification, determining whether or not a given CN statement meets professional standards.

#### *Transformer-based Classifier*

To maximize accuracy, we trained separate classifiers for each of the three evaluation questions in Study 1 (wood stain).[2] The classifiers for "Is Customer Need" and "Sufficient Detail" are based on the output CNs in isolation. The "Follows from the Source" classifier assesses sentence-CN pairs. We benchmarked two foundational LLMs (Claude Sonnet 3.7 and ChatGPT 4o) and three pretrained transformer models (ModernBERT, RoBERTa, and BERT), and found that RoBERTa optimized with varied asymmetric costs and focal loss, yielded the best cross-validation performance across all questions. The superior performance of fine-tuned classifiers relative to LLMs when sufficient labeled data are available is consistent with existing literature (Bucher and Martini 2024; Chae and Davidson 2025; Zhang et al. 2025).

Figure 5 demonstrates that the classifiers effectively replicate professional evaluators. (Recall that the error bars are confidence intervals, not standard errors.) Using an 80/20 cross-validation split of professional-analyst and foundational LLM data from Study 1, the out-of-sample predictions show no statistically significant differences from human evaluations across the nine metrics (using Bonferroni corrections). Notably, the classifiers accurately predict performance for the SFT LLM outputs, despite <u>all</u> SFT LLM examples being excluded from the classifiers' training. These results confirm that the classifiers are reliable proxies for extracting qualitative insights.

[2] The classifiers are trained for a specific task and are accurate for that task. Unlike LLMs, classifiers trained on wood stain do not necessarily replicate professional evaluators in other categories.

**Figure 5**. Evaluating Transformer-based Classifiers for Questions in Study 1

Is Customer Need
Sufficient Detail
Follows from the Source
Percent "Yes" Responses
100%
80%
60%
40%
20%
0%
Professional Analysts
Foundational LLM
SFT LLM
Human Experts (Study 1)
Classifier (Out-of-Sample)

*Impact of Training Sample Size on Fine-tuning*

Figure 6 reports the classifier-predicted performance for the "Is Customer Need" metric as the number of fine-tuning examples varies. Results for "Sufficient Detail" and "Follows from the Source" are available in Appendix F2. We find that performance improves steadily as the training set grows from 75 to 600 examples. However, marginal gains diminish beyond this point. Increasing the training set size above 600 examples does not further enhance the SFT LLM's performance. A similar plateau occurs for the other two metrics, where 600 examples are sufficient to achieve near-optimal results.

**Figure 6**. Predicted Performance with Different Amounts of Fine-tuning Data

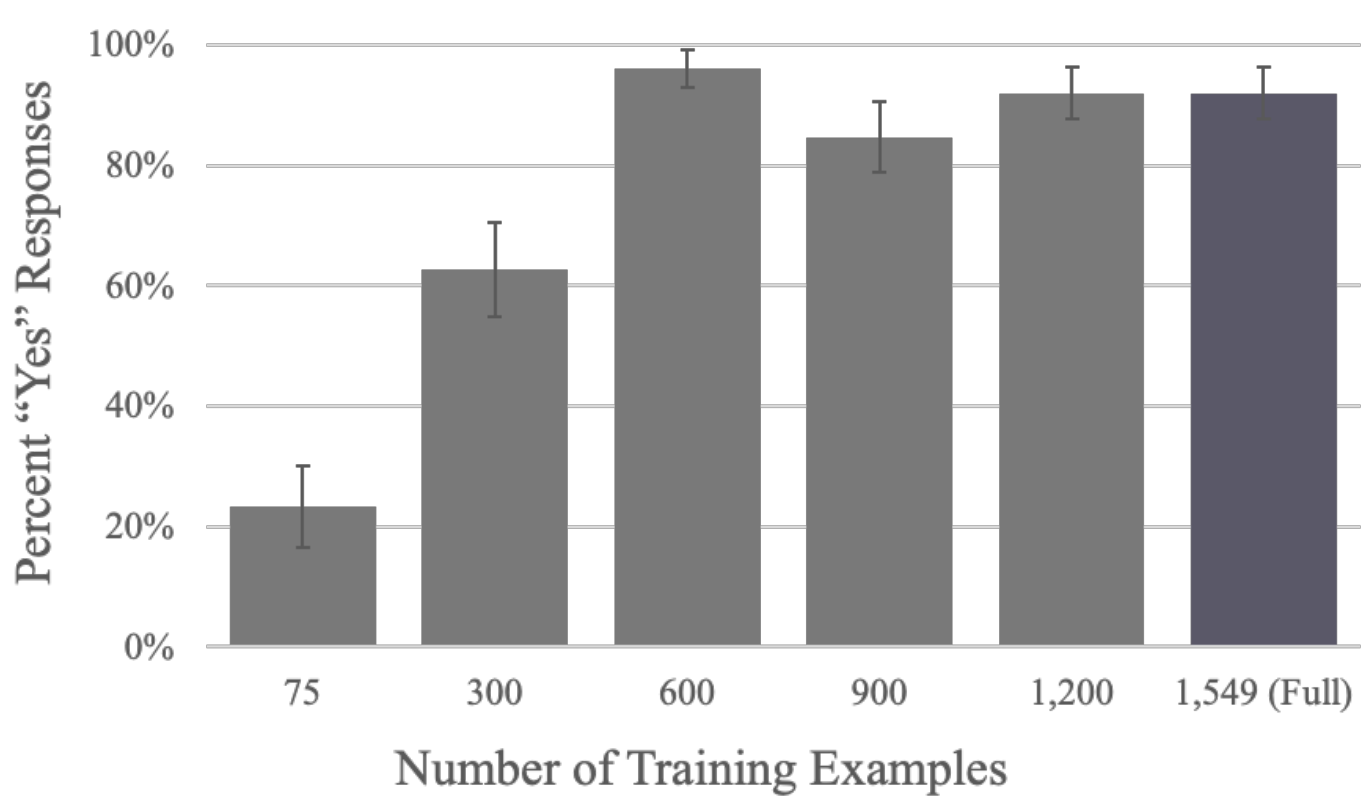

These findings are encouraging for practitioners, particularly for smaller firms with limited resources. While high-quality, professionally abstracted CNs are expensive and difficult to obtain, these results suggest that an SFT LLM can achieve high performance with substantially fewer than 1,549 training pairs. Furthermore, our analysis of category diversity (Appendix F2) indicates that the model is robust to the breadth of training data: performance remained consistent whether the 600 examples were drawn from all ten product categories or a subset of only four.

*Model Size and Generalization Across Foundational LLMs*

Figure 7 reports classifier-predicted performance on "Is Customer Need" as we vary (a) the number of parameters in the foundational LLM and (b) the foundational LLM itself. The results for "Sufficient Detail" and "Follows from the Source" are reported in Appendix F3. To evaluate the impact of model size, we fine-tune the Qwen3 series, an open-weight model family that offers a range of parameter counts with a shared architecture. We focus on the models with 0.6B, 1.7B, 4B, 8B, and 14B parameters. To assess generalizability to different foundational LLMs, we compare Qwen3-14B with DeepSeek-V2-Lite, which represented the state-of-the-art across popular benchmarks at the time of this study. Vicuna-13B is included in both analyses as a baseline.

**Figure 7**. Predicted Performance of Different Foundational LLM Architectures

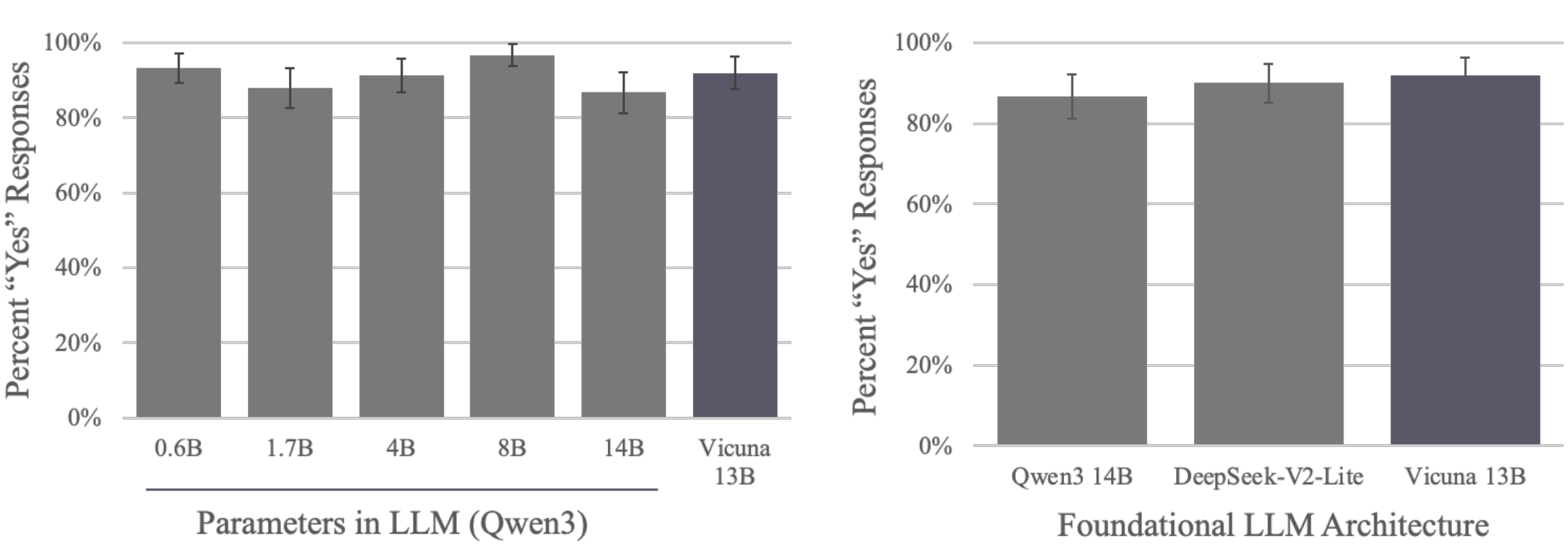

The first panel of Figure 7 indicates that increasing the number of parameters in the foundational LLM does not substantially improve performance after fine-tuning. For example, the Qwen3-0.6B model achieves results comparable to the 14B version and the Vicuna-13B baseline. The second panel suggests that our findings are not unique to Vicuna. Both Qwen and DeepSeek were released one and a half years after Vicuna and, with fine-tuning, appear to perform just as well.

The results in Figures 6 and 7 (and the appendices) provide evidence regarding the fundamental complexity of identifying CNs. On one hand, identifying nuanced CNs is inherently challenging; customers rarely articulate their needs explicitly, so current practice relies on professional analysts with extensive training to capture the underlying "jobs to be done." On the other hand, from a computational perspective, the task involves short-form abstractive summarization of limited text segments, which may not require massive reasoning capacity. Our findings demonstrate empirically that this complex abstraction problem is highly feasible even for small-scale LLMs fine-tuned on relatively modest datasets.

*Evaluating Alternative Prompting Strategies*

Prior to Study 1, we pretested a wide variety of prompting methods. Using the classifiers, we evaluate alternative prompting more formally. Figure 8 reports classifier-predicted performance for the "Is Customer Need" metric as we vary prompting approaches. The zero-shot prompt includes a formal definition of a CN, detailed instructions for formulation (such as requesting concise language and contrasting CNs with solutions), and a defined output format. The simple few-shot prompt removes these detailed instructions but adds 10 CN examples from various product categories. Finally, the detailed few-shot prompt combines both the extensive instructions and the examples.

For the wood-stain category, none of these prompting strategies is sufficient for “Is Customer Need.” However, few-shot prompting does well for “Sufficient Detail” and “Follows from the Source.” See Appendix F4. While increasing LLM context windows may allow for more examples to further improve few-shot performance, we caution that prompting remains fundamentally susceptible to foundational model updates and model drift (Chen et al. 2024). In contrast, fine-tuning allows LLMs to perform on par with professional analysts, providing a stable and reliable solution. This is a critical distinction for practical applications. The primary bottleneck is not the computational cost of SFT, but rather the reliability of the generated customer insights.

**Figure 8**. Predicted Performance of Alternative Prompting Strategies

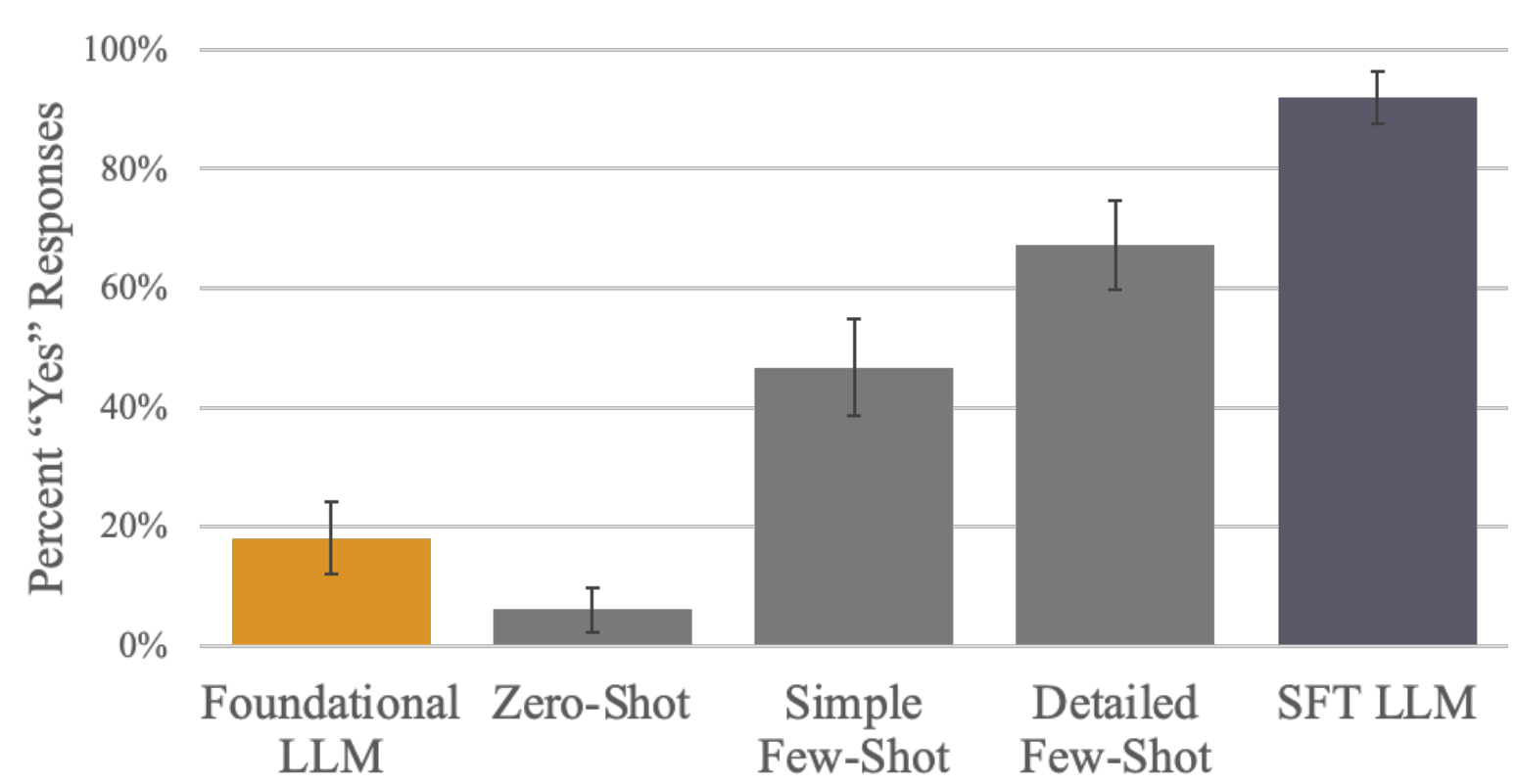

***Study 3. Towards a Deeper Understanding of the SFT LLM Performance (Under the Hood)***

*Transformation versus memorization*

Figures 6 and 7 suggest that CN abstraction is a specialized task requiring a moderate number of fine-tuning examples and a moderately-sized model. In addition, neither the foundational LLM nor the SFT LLM can articulate CNs with sufficient nuance without source material. Together, these observations suggest that the SFT LLM is not storing CNs for retrieval, but is instead learning to transform source material into a specific format. The SFT LLM does not appear to “learn” the customer’s perspective on wood stain as a knowledge base; it learns a process. While the foundational LLM likely contains knowledge of wood stain, it does not seem capable (without fine-tuning or specialized prompting) of inductive out-of-context learning to articulate wood-stain CNs with sufficient breadth, detail, and semantic/syntactic correctness (Treutlein et al. 2024).

To further examine memorization and transformation, we selected source material that required the LLM to draw on stored general knowledge. Consider a well-known closing line from Dickens’ (1859) A Tale of Two Cities: “*It is a far, far better thing that I do, than I have ever done; it is a far, far better rest that I go to than I have ever known*.” This quote and its scholarly analyses are almost certainly present in the training data for the foundational LLM. Not surprisingly, when the foundational LLM is prompted for CNs, it recognizes the quote and suggests the CNs of *meaning, purpose, redemption, and closure*.

In contrast, the SFT LLM (with the CN prompt) focuses on the task it was trained to do. Without the broader context, the SFT LLM returns the abstracted CN: “*a better rest*.” This response is consistent with a task-specific inductive task: a semantic shift from the source material to an abstracted CN, and a syntactic alignment to the preferred format (Ouyang et al., 2022). We found this pattern to hold for other quotes that were likely in foundational training. For example, given Dickens’ opening line, “*It was the best of times, it was the worst of times*,” the SFT LLM ignores the literary context and returns []. These examples suggest that the SFT LLM functions as a transformer of text rather than a retriever of general knowledge.

*Does the SFT LLM Develop a World Model of CN Abstraction*

To investigate whether the SFT LLM has developed a “world model” about the rules for CN abstraction (Mitchell 2025; Rai et al 2024; Vafa et al. 2024), we examine its understanding of the VOC approach. See questions in Table 3. The first question poses a reasoning challenge, which the foundational LLM answers quite well. It correctly parses the prompt and arrives at a valid answer. In contrast, the SFT LLM fails to parse the prompt's intent. Instead, it returns a CN that adheres to its learned syntax.

The second question examines conceptual understanding of the CN-abstraction task. The foundational LLM's world model includes the rules for CN abstraction. The SFT LLM, which is quite good at CN-abstraction, has "forgotten" the rules for CN abstraction (consistent with Luo et al. 2025). Rather than explaining the rules, the SFT LLM simply executes a syntactic and semantic transformation of the prompt into a valid CN. The two diagnostic questions (and others) suggest that fine-tuning does not build a conceptual world model; rather, it teaches the LLM to "paraphrase" source material into the specific structure of customer needs. These insights do not depend upon the foundational LLM that was fine-tuned.

**Table 3**. Questions to Examine the SFT LLM's Understanding of CN Abstraction

| Question | Foundational LLM | SFT LLM |
|---|---|---|
| A team identified 24 distinct customer needs from shaving product reviews and decided to prioritize the top 1/3 of these needs for the next product update. How many needs did they prioritize? | To find out how many needs were prioritized, you need to divide the total number of needs by 3 […] Therefore, the team prioritized 8 of the 24 needs for the next product update. | Able to prioritize the most important customer needs |
| When formulating a need, why is it important to look beyond the specific solution a customer suggests? | By looking beyond the specific solution, you can gain a deeper understanding of the customer's needs and identify opportunities to provide a more comprehensive and effective solution. […] | Able to look beyond the specific solution a customer suggests |

*Failure Modes*

Fine-tuning often induces a new domain-specific language through syntactic and semantic constraints that build upon the representational language building blocks in a foundational model (Ouyang et al. 2022; Treutlein et al. 2024). To examine whether fine-tuning imparts syntactic and semantic constraints, we manually examine the errors made by the foundational LLM and the SFT LLM when formulating CNs. As shown in Figure 9, the SFT LLM outperforms the foundational model across all failure modes. Specifically, fine-tuning effectively taught the SFT LLM to adhere to professional syntactic constraints by avoiding vague goals (generic objectives lacking context) and compound statements (bundling multiple distinct needs). Fine-tuning also instilled semantic constraints, such as avoiding solutions or features (technical implementations), processes or tasks (user actions rather than end states), and opinions or irrelevant text (subjective stories or non-sequiturs). Our research partner notes that these skills are similarly learnable and transferable for human analysts. Once mastered through training, the learned skills apply consistently across product and service categories.

**Figure 9**. Failure Modes for the Foundational LLM and the SFT LLM

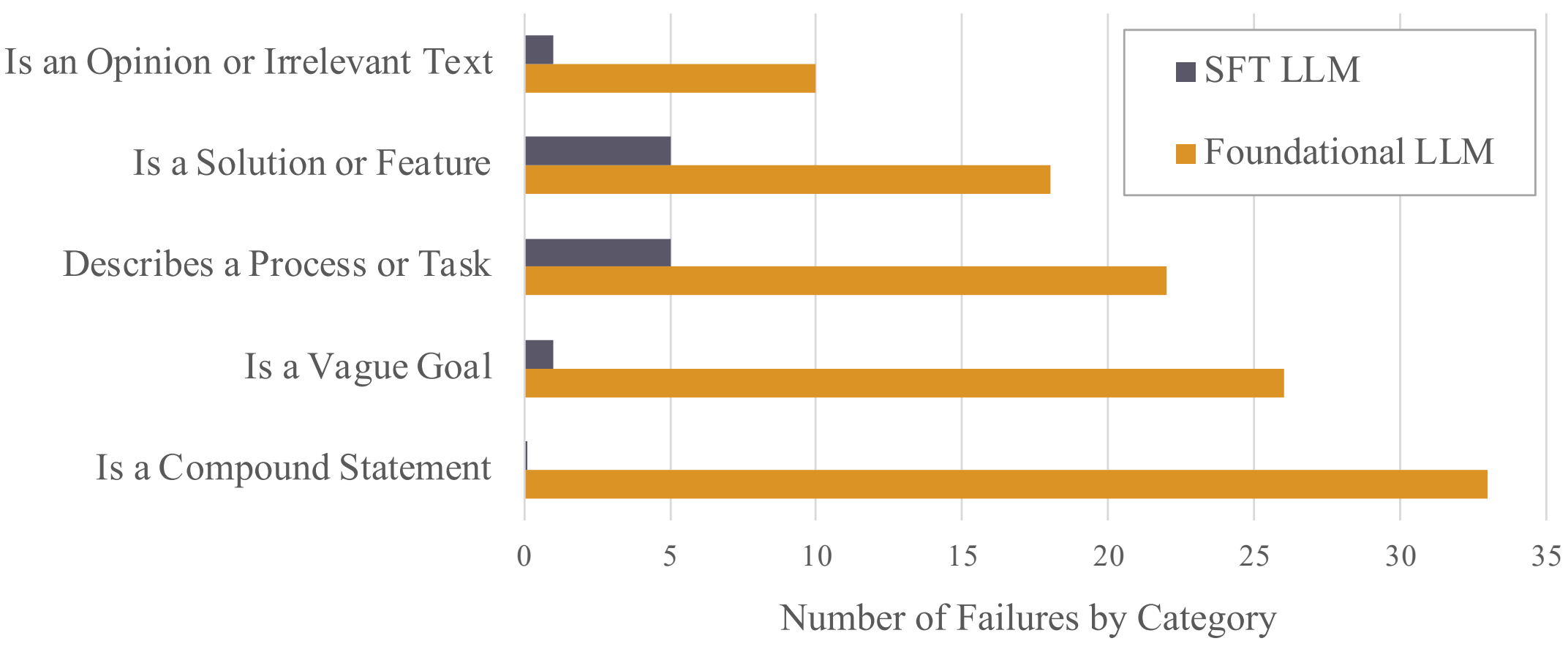

*Summary of Key Findings from Studies 1-3*

Fine-tuning imparts to the SFT LLM a task-specific inductive goal and a new domain-specific language. Rather than relying on stored knowledge for out-of-context learning or product-specific data, the SFT LLMs learn to transform source material into nuanced CNs that strictly adhere to professional syntactic and semantic constraints. The SFT LLMs do not appear to maintain a general “world model” of CNs for specific product categories. They function as a dedicated “CN machine” optimized specifically for abstracting CNs from source sentences.

Figures 3 and 4 demonstrate that, at least for wood stain, this “CN machine” is efficient and effective. Not only does the SFT LLM automate the tedious, labor-intensive task of CN abstraction, but it scales well, enabling firms to identify more high-leverage needs. Facilitated by LLMs, VOC analyses are faster and more effective. Furthermore, because these performance gains are consistent across various foundational models (including smaller architectures) and require only modest amounts of high-quality training data, the power of SFT LLMs can be democratized for innovators who lack the vast resources of larger firms.

Finally, while the SFT LLM “forgets” general-purpose functionality (Luo et al. 2025), this specialization is a feature, not a bug. The SFT LLM is highly proficient at its intended task, while the foundational LLM remains a superior and more viable option for general questions.

## Empirical Evaluation in Other Product Categories

In this section, we examine the performance of an SFT LLM in other product and service categories. We present findings from additional studies in oral care (Study 4) and professional services (Study 5) and summarize insights from four professional applications.

### *Study 4. Oral Care*

We use data from TH 2019 to evaluate the SFT LLM’s performance in the oral care category, following similar blind-test procedures as in Study 1a. The oral care dataset differs methodologically from the wood stain data. TH2019 began with 86 pre-identified CNs and tasked professional analysts with mapping them to a random sample of 8,000 Amazon review sentences. This CN-to-sentence task differs from the wood-stain sentence-to-CN task, thus affecting the professional-analyst benchmark.[3] Nonetheless, we can use the oral-care data to examine the ability of the SFT LLM to abstract CNs. The differing methods provide a convergent test of the basic questions.

Figure 10 compares the performance of professional analysts, the foundational LLM, and the SFT LLM for oral care. The foundational LLM and SFT LLM remain the same, with no additional fine-tuning. Despite the differences between the wood-stain and oral-care studies, the SFT LLM yields a remarkably similar level of performance in absolute terms to its performance in the wood-stain study, suggesting a high degree of stability across product domains.

**Figure 10**. Convergent test of Professional Analysts and LLMs in Oral Care

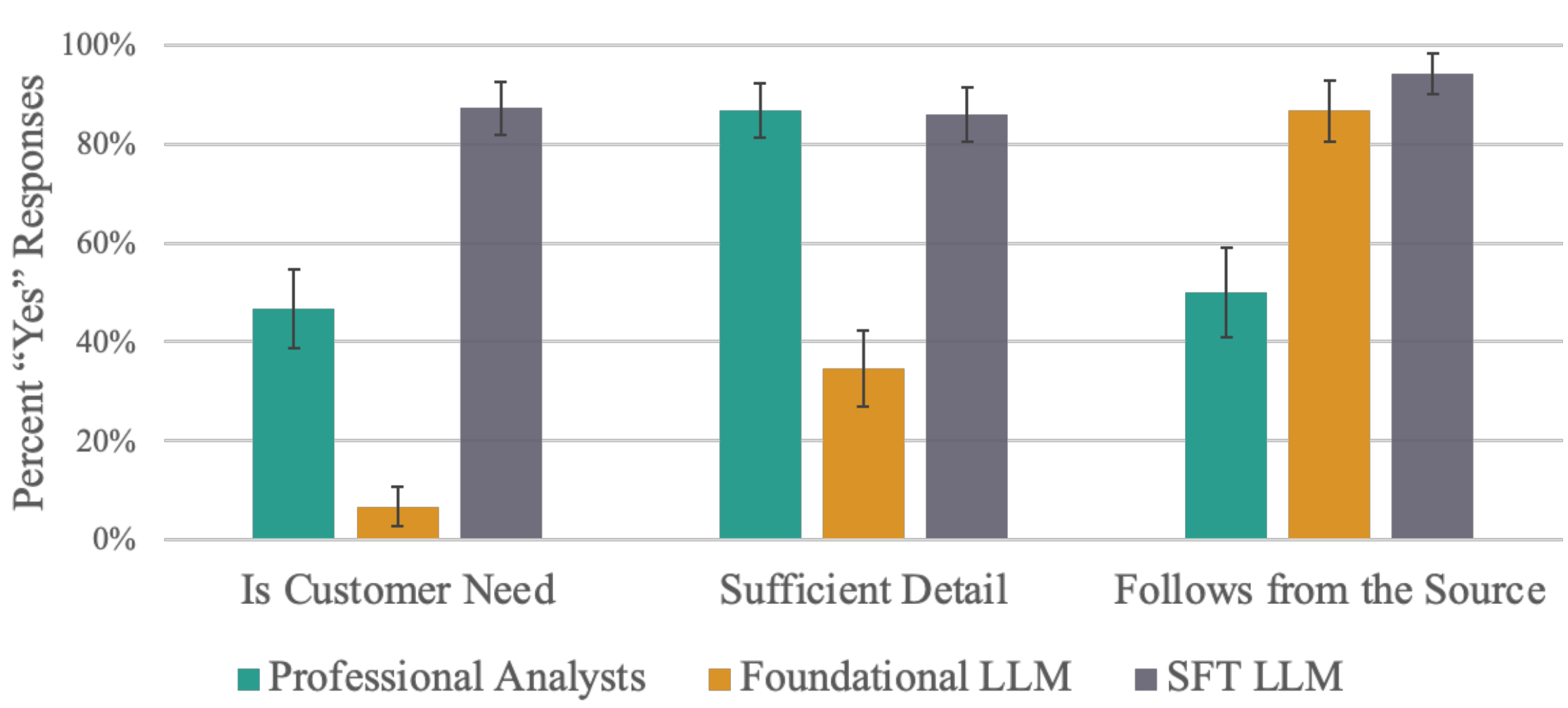

For oral care products, the SFT LLM is again at least as effective as professional analysts. It does substantially better on "Is Customer Need" ($z = 7.83, p < 0.01, h = 0.98$), is comparable on "Sufficient Detail" ($z = 0.18, p < 0.86\ h = 0.02$), and is substantially better on "Follows from the Source" ($z = 7.63, p < 0.01, h = 1.08$). The professional analyst benchmark does less well in oral care than wood stains, likely due to differences in the way in which the professional

[3] The oral-care study predates the wood-stain application; we cannot rule out that analyst training or the mapping task itself influenced the baseline.

analyst benchmark was created (CN-to-sentence vs. sentence-to-CN for the benchmark).

The SFT LLM is substantially and significantly better than the foundational LLM on "Is Customer Need" ($z = 13.99$, $p < 0.01$, $h = 1.89$) and "Sufficient Detail" ($z = 9.08$, $p < 0.01$, $h = 1.11$), and it is slightly better on "Follows from the Source" ($z = 1.98$, $p = 0.49$, $h = 0.26$).

This convergent test in oral care is consistent with Study 1. The SFT LLM, but not the foundational LLM, is managerially and strategically relevant for automating the tedious task of abstracting CNs from source material.

***Study 5. Product Development and Management Association (a Service Category)***

Studies 1-4 are based on product categories and user-generated content. We now examine whether the SFT LLMs extend to a service category and to experiential interviews (as in GH93). Our research partner was asked by the Product Development and Management Association (PDMA) to improve the PDMA's membership services. The PDMA is the premier professional organization for product development, much like the American Marketing Association (AMA) is for marketing, although the PDMA is more practitioner-oriented. The PDMA sponsors national conferences, a journal, local chapters, and provides job-search services. A key service and source of revenue for the PDMA is certification as a new-product-development professional. In cooperation with the PDMA's request, our research partner completed twenty experiential interviews with current and potential PDMA members.

Professional analysts identified 512 CNs, and an SFT LLM identified 1,153 CNs from the interview transcripts. Our research partner used a Qwen SFT LLM due to license restrictions on the Vicuna SFT LLM. Winnowing reduced the CNs to 97 analyst-identified CNs and 124 SFT-LLM identified CNs. The CNs were further winnowed to a merged set of 143 tertiary CNs in the affinity diagram. The SFT LLM identified 100% of the primary and secondary CNs. The PDMA considered the primary and secondary CNs as most critical for strategic examples, while tertiary CNs were used as examples to better understand context. The SFT LLM identified 80% of the tertiary CNs. Tertiary CNs identified by professional analysts, but not the SFT LLM, were close to, but did not duplicate, the included tertiary CNs. As judged by our research partner, the professional-analyst-only CNs had little impact on the final managerial recommendations, and there was nothing systematic. The SFT LLM tertiary CNs were judged by the PDMA and our research partner to be better at identifying niche CNs – an important characteristic given the heterogeneity in the PDMA's customers.

The final CNs provided fundamental insights that changed the way the PDMA operated. For example, the CNs identified that veteran members knew one another and spoke in acronyms causing newcomers to feel like outsiders. Potential members expressed CNs for understandable communication and for rapid integration. Other CNs included CNs for personal gain in careers. Members expressed CNs for ideas and concepts to help them get promoted, to get more respect in their fields, and to improve their performance as product-development professionals. Many CNs were personal, custom, and self-serving. The PDMA knew that community-based CNs among potential members were important, but the identified CNs suggested a different framing of "community." The PDMA responded with new formats for formal events, methods to enhance employment opportunities, new member integration, a broad shift in the roles of members, and a way to facilitate inter-member communication. Many of these changes were introduced successfully at their premier annual meeting. Full details are proprietary and beyond the scope of this paper.

### *Other Applications*

We briefly report four professional applications to the extent that we have received permission. (These are a subset of professional applications.)

*Food brand*. This nationally advertised food brand effectively positioned itself as providing excellent taste, freshness, a visually pleasing presentation, and valuable nutrients. The SFT LLM identified at least one set of new CNs – the customer desired the health value of the nutrients, not the nutrients *per se*. The product could be repositioned to improve a customer's hair, nails, joints, or other health and cosmetic outcomes.

*Top-ranked specialized hospital*. This well-known hospital provided complex care via multiple medical specialists. High-quality care to its targeted customers is a "must have" CN. But by identifying and satisfying the family's CNs throughout the journey from specialist to specialist, the hospital achieved greater satisfaction among all of its customers.

*Composite siding*. Based on 32 transcripts from concept-testing interviews with homeowners (B2C), architects, and contractors (B2B), the SFT LLM abstracted CNs to provide a comprehensive set of 3,901 CNs ultimately winnowed to 103 tertiary CNs. As in Studies 1, 4, and 5, the CNs identified by the SFT LLM were comparable to those identified by professional analysts.

*B2B paint and stain*. While Studies 1, 2, and 3 address B2C CNs for wood stain, this application demonstrated the feasibility for B2B applications based on the data from professional forums. New insights were obtained.

### *Qualitative Observations Based on Applications to Date*

Formal evaluation of the effect of artificial intelligence on the role of professional analysts is an important research issue beyond the scope of this paper (e.g., Brynjolfsson, Chandar, and Chen 2025; Challapally et al. 2025). Nonetheless, to foster further research we provide qualitative observations based on the applications we have observed to date.

The SFT LLM is changing the role of a professional analyst. Professional analysts still interview customers or assemble user-generated content, but once analysts identify source sentences, the SFT LLM automates the tedious task of abstracting CNs. Analysts winnow the CNs directly because CNs are in the right syntax and are similar semantically. Winnowing is more efficient and VOC analyses scale to much larger corpora leading to a more-comprehensive set of tertiary CNs. VOC studies are more effective and lower cost and within the reach of a broader set of firms, especially entrepreneurs. At our research partner, each analyst is now more efficient and more valuable to the firm, and professional tasks are more rewarding and enjoyable.

## Summary, Limitations, and Future Research

### *What we learned from Studies 1 through 5*

Across five interrelated studies, we provide convergent evidence that SFT LLMs perform on par with professional analysts in identifying customer needs. *A priori,* the need for supervised fine-tuning was not obvious. Given that LLMs excel at many unstructured, human-like tasks, it would not have been surprising if a foundational LLM performed well. However, we found that untrained foundational LLMs do significantly and substantially worse on CN abstraction. While few-shot prompting shows promise, fine-tuning is quite feasible and is likely the better and more reliable option.

The fine-tuning process effectively mimics how human professionals are trained and teaches the LLM the specific syntactic and semantic constraints of CNs. The SFT LLM does not rely on memorization. Rather, it becomes a specialized processor that transforms almost any input into the learned CN format, often at the expense of general-purpose reasoning. Because CN abstraction is a highly specialized task focused on short-form transformation, smaller-parameter models fine-tuned with relatively few examples appear sufficient to achieve professional-grade results.

### *Challenges Yet to be Met*

While SFT LLMs appear up to the job of abstracting CNs, there are many steps in the VOC that remain human-centric. Winnowing abstracted CNs is faster and easier than winnowing raw source sentences. LLMs facilitate this process, but the validation of LLM-generated affinity diagrams remains an open research question. Historically, clustering methods such as embeddings of source content, topic models, and keyword searches have struggled to provide actionable structures. We are optimistic that the standardization inherent in SFT-generated CNs may finally enable machine-based clustering to yield effective automated affinitization.

SFT LLMs can abstract CNs but cannot yet prioritize them. Researchers have attempted to use frequency, star-labeled ratings, and sentiment as indicators of importance, but such indicators are at best weakly correlated with CN importance and at worst counterproductive (GH 1993, TH 2019). Prioritization by machine learning remains elusive.

As we automate more components of the VOC workflow, we envision a future where both the emergence and fulfillment of CNs are monitored automatically and continuously to enhance the management of products, services, or brands. There are many milestones between the current state-of-the-art and this blue-sky goal, but automating CN abstraction is a valuable step along that path.

# Appendices

## Appendix A. Ten Product Categories Used for Model Training

- Activewear
- Glucose Monitoring
- Hearing Aids
- Housing and Apartments
- Lawncare Equipment
- Men’s Shaving
- Sleep Aids
- Snow Removal Equipment
- Telehealth
- Women’s Underwear

## Appendix B. Vicuna Fine-tuning and Inference

We fine-tuned Vicuna 13B using 4x Nvidia A100 PCIe on Lambda GPU Cloud. The system provides 40 GB VRAM, 120 vCPUs, 800 GB RAM, and 1 TB SSD storage. The fine-tuning process used bf16 precision without quantization, running for 6 epochs with a per-device batch size of 2 for training and 8 for evaluation. It employed a gradient accumulation of 4, a learning rate of 2e-5, cosine scheduling, and a maximum sequence length of 1024. Fine-tuning took approximately 8 hours. Inference, performed on 1x Nvidia A100 PCIe, takes about 0.4–0.5 seconds per prompt, or 400–500 seconds per 1,000 prompts.

## Appendix C. Additional Examples of Customer Needs for Wood Stain Products

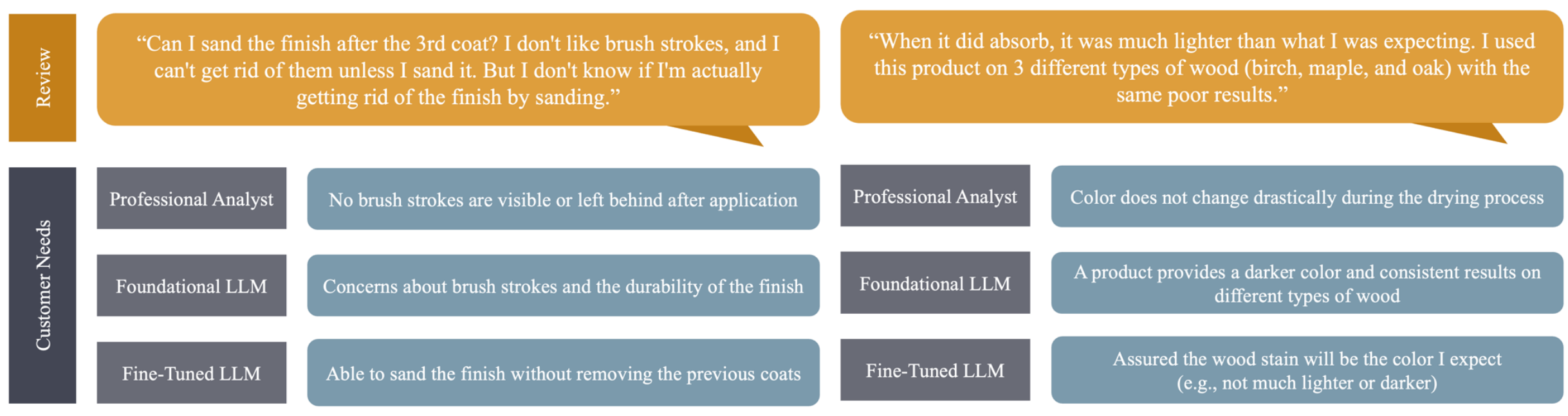

In Panel A, the professional analyst captures a CN perfectly, which demonstrates a deep understanding of the “job to be done.” The SFT LLM abstracts a different CN. This CN is real, but it does not answer the question: Why do customers want to sand after the finish? The Base LLM reformulates generic concerns, instead of articulating a CN.

In Panel B, the professional formulation is a meaningful CN, but the idea differs from the original review. The Base LLM abstraction is a statement that does not look like a professional CN and misrepresents the information from the review. The SFT LLM yields the desired result.

## Appendix D. Detailed Instructions in Blind Studies

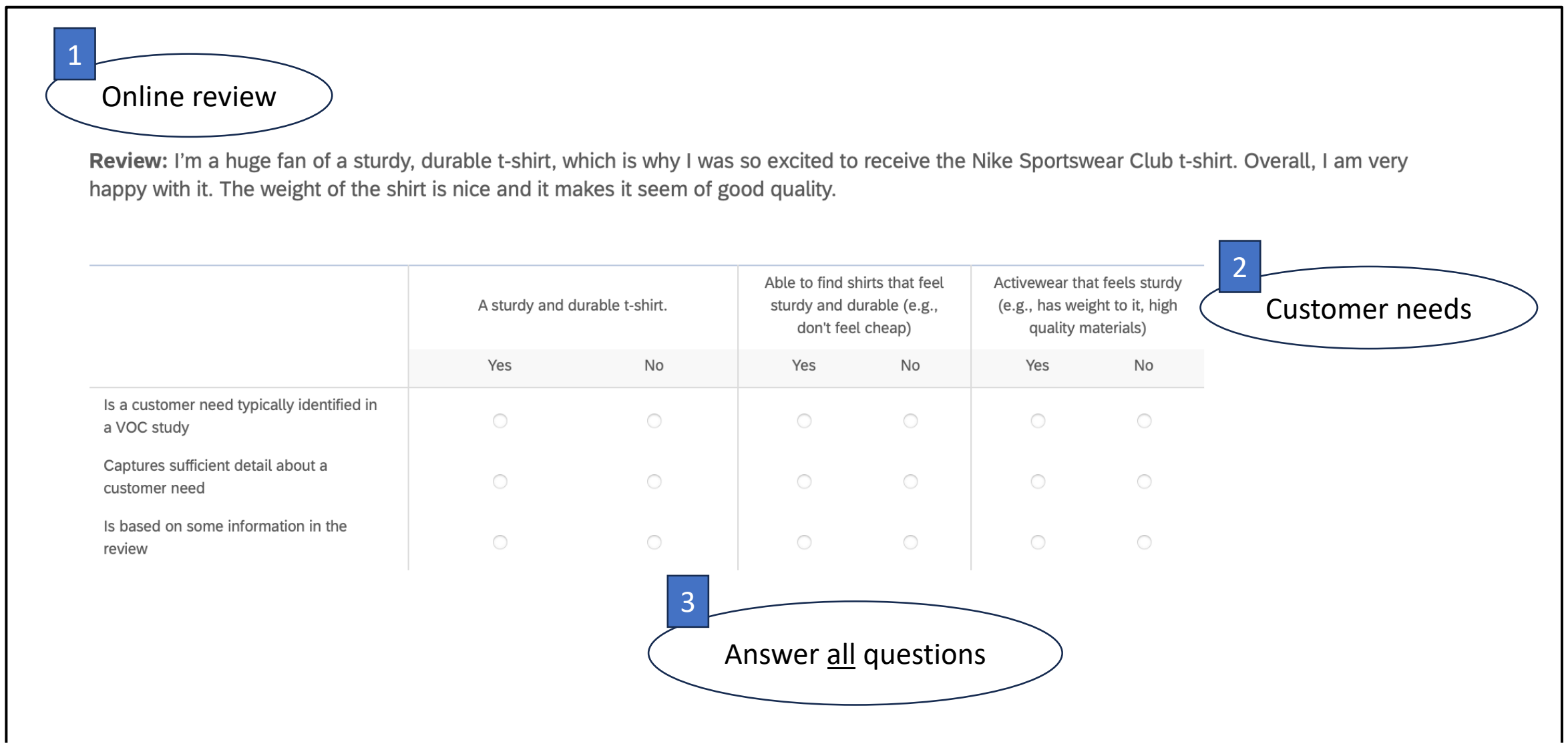

### Q1. Is a customer need typically identified in a VOC study.

Please indicate whether the statement qualifies as a customer need identified in a typical VOC study. Customer needs capture conceptual benefits that customers want to obtain from a product, which is different from customer-provided technical specifications and desired solutions.

*General Comment:* For Q1, evaluate only if the statement is a customer need, regardless of whether the statement is detailed enough, which will be judged in Q2. This question also does not evaluate whether the statement came from the review, which will be judged in Q3.

### Q2. Captures sufficient detail about a customer need.

Please evaluate whether or not the statement is actionable and not too general. For example, "good communication" might be too general. "Can stay informed of the technician's status (e.g. when they will arrive)" captures sufficient detail.

### Q3. Is based on some information in the review.

Please evaluate whether or not the statement is based on information in the review. In particular, is it reasonable that a VOC study would abstract this customer need from the review?

## Appendix E. Results for Study 1 Disaggregated by the Type of Source Content

| | **Verbatim** | | | **Informative** | | | **Uninformative** | | |
|---|---|---|---|---|---|---|---|---|---|
| | Prof. Analyst | Found. LLM | SFT LLM | Prof. Analyst | Found. LLM | SFT LLM | Prof. Analyst | Found. LLM | SFT LLM |
| Is Customer Need | 0.81 | 0.33 | 0.92 | 0.80 | 0.20 | 0.90 | 0.93 | 0.17 | 0.93 |
| | (0.04) | (0.05) | (0.03) | (0.07) | (0.07) | (0.06) | (0.05) | (0.07) | (0.05) |
| Sufficient Detail | 0.94 | 0.73 | 0.89 | 0.83 | 0.73 | 0.97 | 0.87 | 0.73 | 0.83 |
| | (0.02) | (0.05) | (0.03) | (0.07) | (0.08) | (0.03) | (0.06) | (0.08) | (0.07) |
| Follows from the Source | 0.91 | 0.90 | 0.98 | 0.03 | 0.93 | 0.93 | 0.00 | 0.53 | 0.87 |
| | (0.03) | (0.03) | (0.02) | (0.03) | (0.03) | (0.05) | (0.00) | (0.09) | (0.06) |

We highlight two observations. First, on the dimension "Follows from the Source," the performance of the professional-analyst baseline drops close to zero for *informative* and *uninformative* reviews. This serves as an attention check in our survey design because we randomly selected random professional-analyst-abstracted CNs for these reviews from the original VOC study.

Second, we observe that the performance of the LLMs on the "Follow from the Source" dimension is lower for *uninformative* reviews than for *informative* reviews and *verbatims*. Recall that professional analysts found these reviews uninformative in the professional VOC study. Our design substituted other CNs, so as expected, evaluators agreed that the analyst-abstracted CNs are indeed CNs. The SFT LLM was able to abstract CNs for many of the analyst-designated *uninformative* reviews, suggesting that the SFT LLM is more efficient than analysts at identifying CNs. The foundational LLM abstracted some CNs, but not as many as the SFT LLM. The foundational LLM shows evidence of hallucination in the "follows from the source" question, particularly for uninformative reviews. The SFT LLM is more robust.

## Appendix F. Study 2 Results for "Sufficient Detail" and "Follows from the Source"

**Figure F1**. Predicted Performance with Different Amounts of Fine-Tuning Data

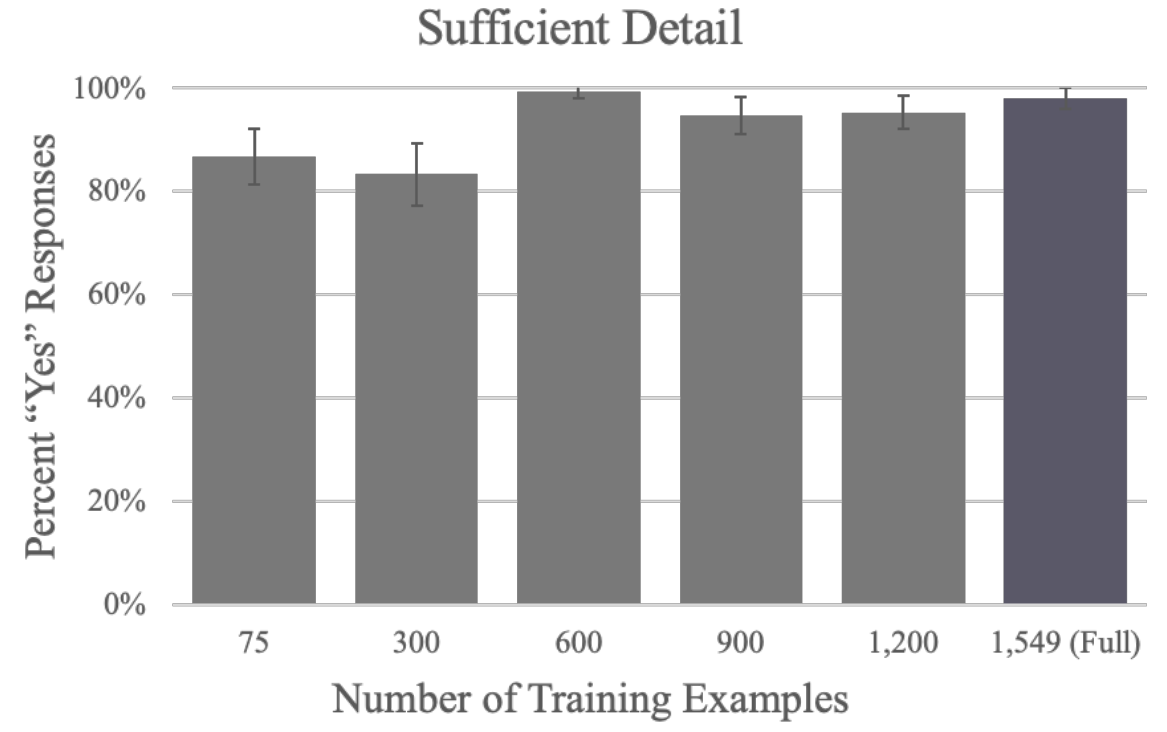

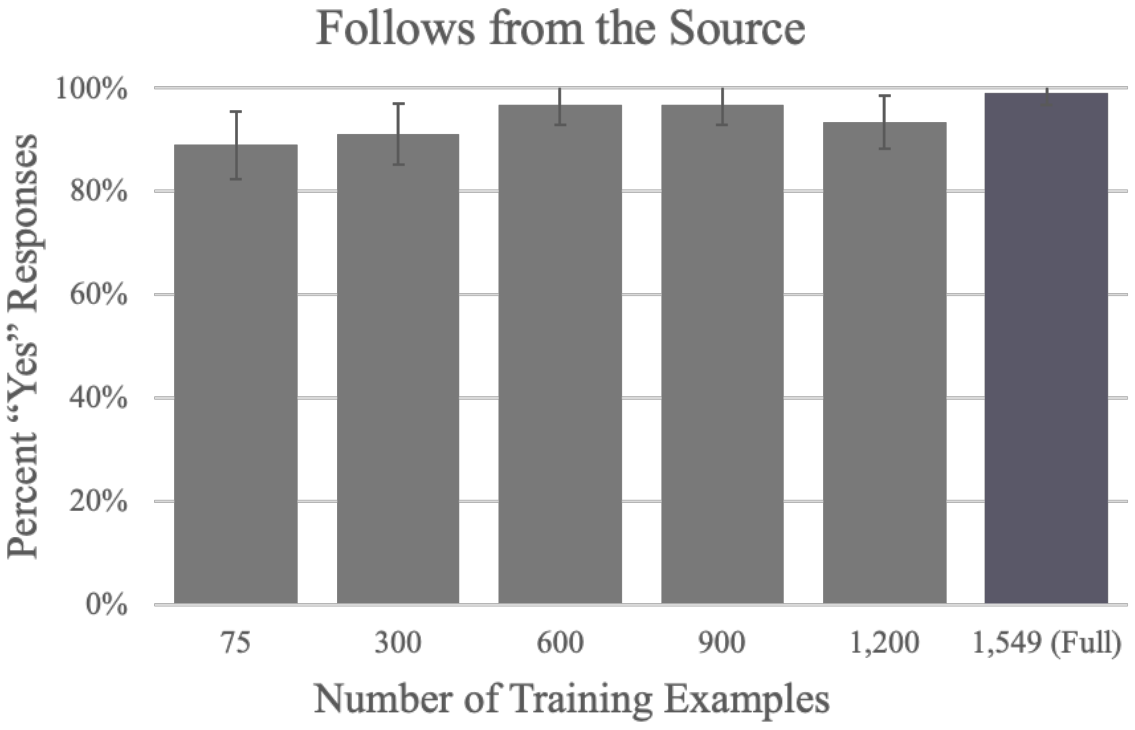

**Figure F2**. Predicted Performance with 600 Training CNs from Different Number of Categories

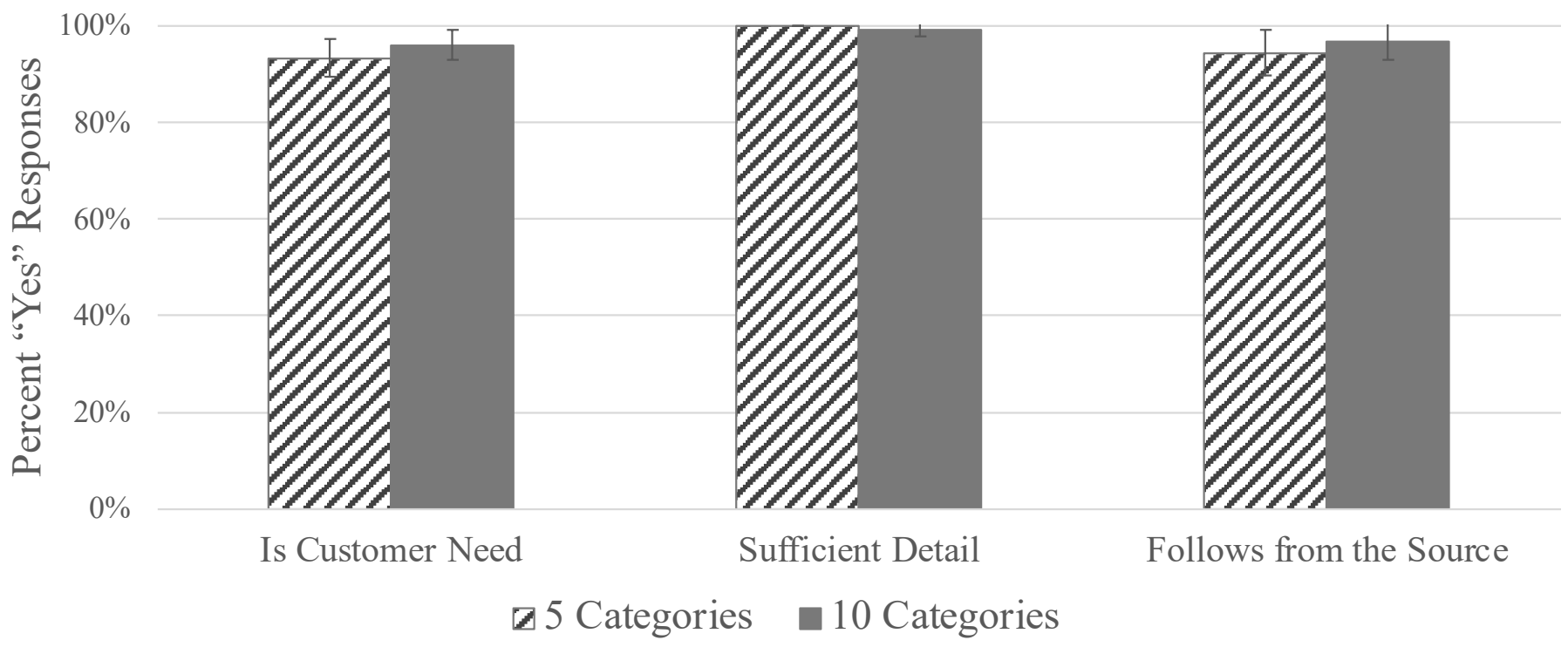

**Figure F3**. Predicted Performance of Different Foundational LLM Architectures

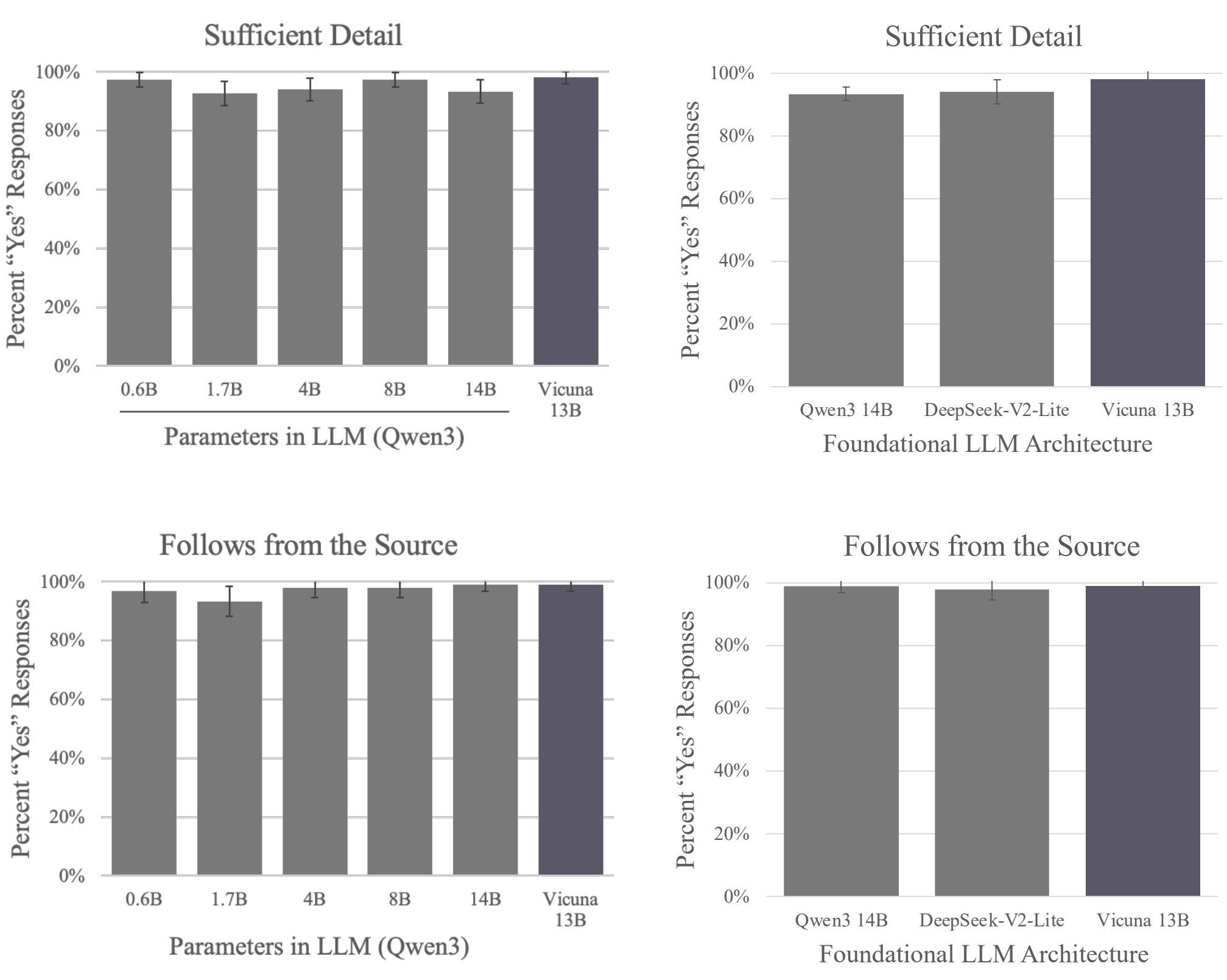

**Figure F4**. Predicted Performance of Alternative Prompting Strategies

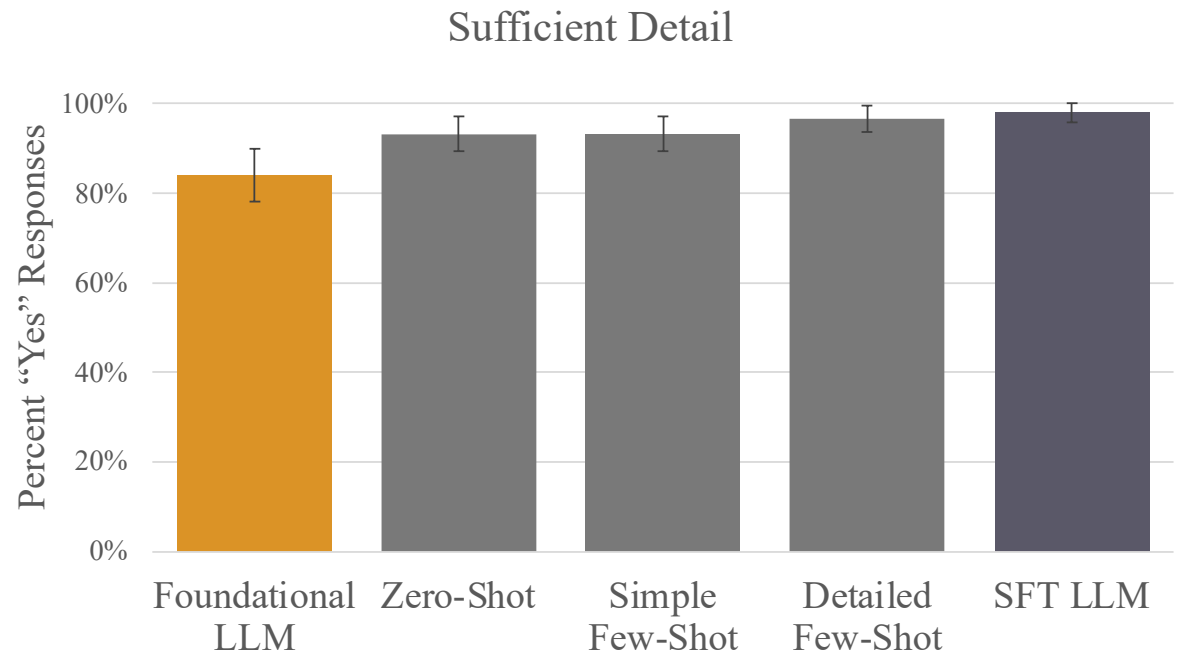

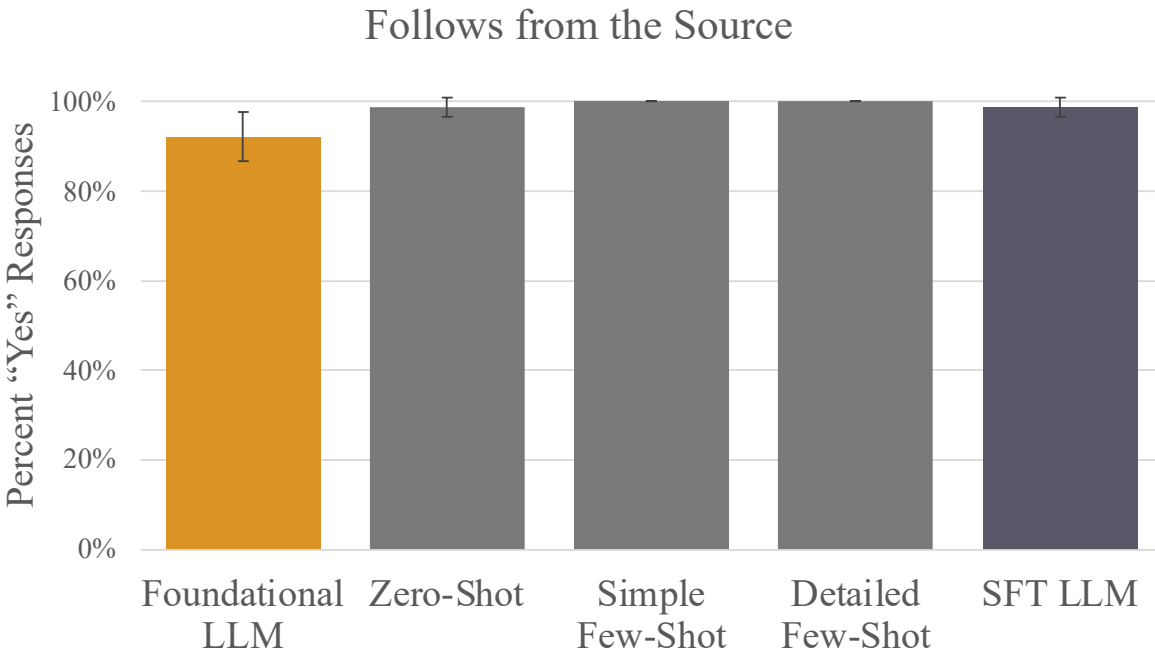